\documentclass{article}

\usepackage{microtype}
\usepackage{graphicx}
\usepackage{subfigure}
\usepackage{booktabs} 

\usepackage{hyperref}

\usepackage[accepted]{icml_fmwild}

\usepackage{amsmath}
\usepackage{amssymb}
\usepackage{mathtools}
\usepackage{amsthm}
\usepackage{enumitem}

\usepackage[capitalize,noabbrev]{cleveref}

\theoremstyle{plain}

\theoremstyle{definition}

\theoremstyle{remark}

\usepackage[textsize=tiny]{todonotes}

\icmltitlerunning{A Greedier Perspective on the Weight-Ensemble}

\begin{document}

\twocolumn[
\icmltitle{Understanding the Role of Functional Diversity in Weight-Ensembling with Ingredient Selection and Multidimensional Scaling}



\begin{icmlauthorlist}
\icmlauthor{Alex Rojas}{yyy}
\icmlauthor{David Alvarez-Melis}{yyy,comp}
\end{icmlauthorlist}

\icmlaffiliation{yyy}{Harvard University, Cambridge, MA, USA}
\icmlaffiliation{comp}{Microsoft Research, Cambridge, MA, USA}

\icmlcorrespondingauthor{Alex Rojas}{alexnrojas5@gmail.com}

\icmlkeywords{Machine Learning, ICML}

\vskip 0.3in
]



\printAffiliationsAndNotice{}  





\begin{abstract}
     Weight-ensembles are formed when the parameters of multiple neural networks are directly averaged into a single model. They have demonstrated generalization capability in-distribution (ID) and out-of-distribution (OOD) which is not completely understood, though they are thought to successfully exploit functional diversity allotted by each distinct model. Given a collection of models, it is also unclear which combination leads to the optimal weight-ensemble; the SOTA is a linear-time ``greedy" method. We introduce two novel weight-ensembling approaches to study the link between performance dynamics and the nature of how each method decides to use apply the functionally diverse components, akin to diversity-encouragement in the prediction-ensemble literature. We develop a visualization tool to explain how each algorithm explores various domains defined via pairwise-distances to further investigate selection and algorithms' convergence. Empirical analyses shed perspectives which reinforce how high-diversity enhances weight-ensembling while qualifying the extent to which diversity alone improves accuracy. We also demonstrate that sampling positionally distinct models can contribute just as meaningfully to improvements in a weight-ensemble.
\end{abstract}

\section{Introduction}
\label{introduction}

Model ensembling plays a crucial role in enhancing the performance and robustness of machine learning models. Combining the information learned by models pre-trained (or fine-tuned) with different configurations or on different tasks can reduce overfitting to any particular hyperparameter or dataset choice, leading to better generalization \cite{lakshminarayanan2017simplescalablepredictiveuncertainty}. This traditional approach to model ensembling in machine learning relies on averaging the \textit{predictions} of the various models, which suffers from high memory and requires the inference-time computational cost of many modern neural networks.


Recent work by \citet{wortsman2022model} demonstrates a promising alternative.  The  ``model-souping" approach directly averages the parameters of a host of models in weight-space. This weight-ensembling operation results in a single weight-average model (the WA), addressing many of the computational limitations of prediction-based ensembling. The highly non-convex neural network loss landscape suggests that this approach should fail; yet, the linear mode connectivity (LMC) property of such landscapes \cite{frankle2020linear} demonstrates that the interpolation between models sharing a stable region of a weight-space training trajectory remains in a low-error region. An example includes when minima have been fine-tuned from a shared foundation model; minima in this setting reside in convex low-loss basins where weight-ensembling does not incur significant loss barriers \cite{neyshabur2021transferred}.

Weight-ensembling literature typically assumes access to multiple models of identical architecture fine-tuned from a shared initialization, though fine-tuning hyperparameters may vary. Prior work relies on a ``greedy" approach to construct the ensemble, whereby models (the ``ingredients'') are considered only once to be sequentially added to the ``soup'' based on validation set accuracy of a candidate WA; ingredients are individually thrown out if candidate performance declines \cite{wortsman2022model,rame2022diverse}. \looseness-1 These greedy WA models have shown remarkable performance across 
complex tasks such as ImageNet \cite{5206848} and DomainBed \cite{gulrajani2020search}, as shown by \citet{wortsman2022model} and \citet{rame2022diverse} respectively.

While the empirical success of weight-ensembling is linked to its discovery of flatter regions of the loss landscape \cite{wortsman2022model, cha2021swad} and its incorporation of functionally diverse ingredients \cite{rame2022diverse}, the mechanics of how the greedy algorithm elicits these phenomena is not clear. Additionally, it is unknown if the greedy algorithm is particularly well-suited to find diverse ingredient-sets, or whether there exists other methods better suited to this goal.  \looseness-1

In this work, we study the link between functional diversity and the performance of various weight-ensembling methods to further understand the effectiveness of weight-ensembling. Grounded by discussion of existing bias-variance-diversity decompositions of prediction-ensemble error and subsequent analysis of weight-ensemble error, we can renegotiate how this bias-variance-diversity tradeoff is understood for weight-ensembling. To that end, we can reason about the utility of promoting several quantitative diversity measures by characterizing how three weight-ensembling algorithms leverage these relationships and perform. Two are novel: ``greedier" serves as a costly, optimal benchmark for comparison to others, while ``ranked" plays the role of a diversity-encouraging mechanism; ``greedier" is adopted from the literature \cite{rame2022diverse}, \cite{wortsman2022model}. In summary, our contributions are:



\begin{itemize}[topsep=1pt, parsep=1pt, itemsep=0pt, partopsep=0pt,leftmargin=*]
\item We develop the ``greedier'' algorithm for weight-ensembling, which has the flexibility at each iteration to add any model to the set of ingredients. Although computationally costly, greedier serves as a benchmark due to its optimality. 
\item We develop the ``ranked'' algorithm for weight-ensembling, which considers ingredients in order of decreasing diversity from the current WA and greedily selects the first candidate which improves performance
\item We empirically show how two notions of diversity robustly explain differences in the selection mechanism of the ``greedier'' and ``greedy'' algorithms. The advantageous performance of greedier implies that the presence of such distances help improve WA accuracy efficiently. However, the most diverse selections made by the diversity-encouraging ``ranked'' algorithm perform less optimally than ``greedier,'' limiting the extent to which selecting for maximal diversity is useful.
\item We introduce a form of qualitative visualization that provide additional insights on the connection between these weight-ensembling algorithms and loss landscapes.

\end{itemize}
 

\section{Related Work}


Traditional prediction-ensembling theory depends on the harmonious combination of distinctive predictive mechanisms to reduce generalization error. Ensemble error was first shown to decompose into the average error of the ensemble members minus the ensemble ambiguity; the ambiguity is a positive value reflecting the variance of ensemble-member predictions around the ensemble prediction to serve as an early quantification of diversity \cite{krogh1994neural}. By maximizing this notion of diversity, pracitioners hoped to reduce generalization error, motivating the next stage of deep ensembling approaches.


Empirical navigations seeking to maximize diversity within this bias-variance-diversity tradeoff are abundant. \citet{lakshminarayanan2017simplescalablepredictiveuncertainty} prediction-ensemble independently initialized and trained neural networks, implicitly leveraging the diversity derived from the stochastics in those steps to reduce generalization error. \citet{fort2020deepensembleslosslandscape} later describe how well-trained samples from two distinct such modes exhibit more diversity than do distinct samples from a single trajectory of training.  Resulting from the strong performance of deep ensembles, approaches to further \textit{explicitly} encourage diversity between ensemble members have ensued. For example, incremental improvement was demonstrated through ensembling over members trained using carefully selected hyperparameter ranges instead of shared hyperparameters \cite{wenzel2020hyperparameter}. Additionally, diversity-encouraging regularization of the loss during the joint-training of ensemble members has been explored were explored in \citet{ortega2022diversity} and \citet{abe2022best}.

Bias-variance-diversity decompositions were generalized for arbitrary loss functions which admit bias-variance decompositions \cite{wood2024unifiedtheorydiversityensemble}. While previously, such a decomposition had motivated the ensembling of diverse low-bias neural networks, \citet{wood2024unifiedtheorydiversityensemble} underscore that much like the traditional bias-variance decomposition, the bias-variance-diversity decomposition must be examined as a tradeoff that must be carefully managed; simply maximizing diversity may have the negative externality of degrading individual-model performance and the ensemble itself by extension.


Recently, \citet{abe2023pathologies} provides a more thorough examination for shallow and deep networks. Deeper architectures trained on a loss that regularizes the joint member task-loss by rewarding diversity have run up against a member-performance member-diversity frontier. In this case, increasing diversity degrades individual member performance by increasing bias and thus adversely impacts the bias-variance-diversity tradeoff both in-distribution and out-of-distribution. 



Given that the weight-ensembling results in one fixed mode for inference, how might error be reduced through the machinations of diversity? \citet{rame2022diverse} prove a bias-variance-diversity tradeoff for weight-ensembles by applying a first-order approximation of the prediction-ensemble with the weight-ensemble -- although this approximation decays under a quadratic locality asymptote. The authors then encourage diversity by fine-tuning models from a shared initialization, varying hyperparameters; the resulting weight-ensembles achieve state-of-the-art at the time on OOD tasks. Maximizing the average pairwise diversity between ingredients is credited for the success of the approach. But, the notion of the average pairwise diversity \textit{between members} is disjoint from diversity of members \textit{from the ensemble prediction}; the latter idea characterizes the ambiguity terms consistent with ensemble theory. Taken together, existing weight-ensemble analysis lacks a careful navigation of the bias-variance-diversity error decomposition as a tradeoff.  With similar motivation to \citet{abe2023pathologies} 's work in prediction-ensembling, our work thus seeks to intricately analyze the balance of the bias-variance-diversity tradeoff in the case of weight-ensembles.



\section{Methodology}

\subsection{Distance Measures Between Models}

Functional diversity, as measured by the ratio-error \cite{aksela2003comparison} was claimed in \citet{rame2022diverse} to be the driving force behind the improvements of weight-ensembling over standard (SGD-found) models. This stemmed from analysis claiming that functional diversity decorrelates model predictions, with the latter being a term residing in a bias-variance-covariance decomposition of generalization error. They also demonstrate a positive correlation of the average pairwise diversity of a set of models with the accuracy gain of the corresponding WA over the mean accuracy over the individual ingredients, a statistic which does not directly imply that there is a lack of functional redundancy in the collection, only in the average case of a pair. For this reason, when analyzing weight-ensembling algorithms iteration-by-iteration in this work, we also pay heed to the diversity between a selected candidate ingredient and the current WA (in the context of unselected ingredients' distances). Less lossily than average pairwise diversity, we analyze these pairwise relationships jointly in the visualization method. 

\textbf{Definition 2.1.} We use the convention of ratio-error \cite{aksela2003comparison} to measure diversity following \citet{rame2022diverse}. For models $\theta_A$ and $\theta_B$, where $N_{uns}$ and $N_{sha}$ refer to the number of unshared and shared errors on a labelled dataset,  we refer to the diversity distance as $d_D (\theta_A, \theta_B) = \frac{N_{uns}}{N_{sha}}$

Motivated by the finding of loss basins in fine-tuning \cite{neyshabur2021transferred}, convex regions in which models sharing initialization remain essentially linearly mode connected, we also use a Euclidean geometry-inspired measure to determine the extent to which specific weight-space geometry can explain weight-ensembling approaches. The Euclidean metric allows us to explore how different weight-ensembling traverse the basin and evaluate whether sampling candidates different parts of a loss basin sufficiently improves the WA, recasting the pursuit diversity as a weight-space traversal question. 



\textbf{Definition 2.2.} As such, we define the Euclidean distance between two neural network parameters $\theta_A, \theta_B \in \Theta$ to be $d_E(\theta_A, \theta_B) = || \textrm{vec}(\theta_A)  - \textrm{vec}(\theta_B) ||_2^2$


\subsection{The Greedy Weight-Ensembling Technique}

The greedy souping method \cite{wortsman2022model, rame2022diverse} sorts the individual models by decreasing validation accuracy. Starting from the single highest-accuracy model, we sequentially consider adding the remaining models, only adding to the ingredients list when a candidate WA improves training-domain validation accuracy and otherwise throwing out the failed candidate in a linear pass.



\subsection{A Greedier Weight-Ensembling Technique}

New to this work, the \textit{greedier} weight-ensembling algorithm also initializes to the top validation-accuracy ingredient. At each step, we consider the inclusion of every remaining ingredient to the set, aggregating the candidate whose WA maximally performed to the set if we have outperformed the current set's WA accuracy. If no candidate set's WA has outperformed the current set WA's accuracy, the algorithm terminates. See Algorithm \ref{alg:greedier} for granular details. The core difference between the algorithms is that instead of the greedy algorithm's one single linear pass through the models sorted by individual performance, the greedier algorithm can add models in any order if they still contribute positively to the soup. The similarity is that both algorithms initialize the ingredients list to the maximal performing model.

While this algorithm has a costly runtime, it serves to illuminate the measures which maximally explain the selection mechanism of the algorithm. This will help diagnose what drives the selection of new ingredients to understand what relationships between ingredients and the current WA contribute to maximal improvements in the WA. Treating greedier as a ``gold-standard" benchmark also allows us to correlate other algorithms' selection mechanisms with their performance characteristics.


\subsection{Ranked Weight-Ensembling}

The next weight-ensembling algorithm introduced here, the \textit{ranked} algorithm, initializes identically to previous algorithms. At each step, we sort remaining ingredients by decreasing distance from the current set's WA and proceed through these rankings considering the addition of each ingredient to the set individually. The first ingredient whose candidate set improves training domain accuracy is accepted, and the rejected models are stashed for the next iteration. See Algorithm \ref{alg:ranked}. The intent of this method is to make salient the effect of biasing diversity into the selection mechanism, which allows us to tease out benefits which stem from the inclusion of diverse candidates and less of the confounding factors which may affect the greedier and greedy algorithms. Furthermore, considering the diversity between the current weight-ensemble and new ingredients, instead of between the ingredients as in \cite{rame2022diverse}, is a more principled parallel to bias-variance-diversity decompositions in which the ambiguity terms reflect variance around ensemble predictions, not between members \cite{wood2024unifiedtheorydiversityensemble}.

\section{Experiments}

\subsection{Experimental Setting}

We adopt the DomainBed setting \cite{gulrajani2020search} used by \citet{rame2022diverse}, honing the focus of our experiments to the OfficeHome dataset \cite{venkateswara2017deep}, a domain generalization dataset containing four test environments. We refer the reader to Section 3.2 of \citet{rame2022diverse} which established the setting adopted here, including the random initialization fine-tuning setup for ResNet50
\cite{he2016deep} trained on ImageNet \cite{5206848} as a foundation model varying hyperparameters and randomness to obtain distinct fine-tuned models. Holding out one test environment at a time as OOD set, we run ten trials of fine-tuning on the three ID environments 40 models per trial. After the fine-tuning, we run the greedier, greedy, and ranked weight-ensembling algorithms. For the ranked method, we run two versions, one which ranks ingredients by ratio-error which we call ``diversity-ranked'' and the other using Euclidean distance called ``Euclidean-ranked.''

\subsection{Weight-Ensemble Algorithm Performance Results}

\begin{figure}[ht]
\vskip -0.1in
\begin{center}
\centerline{\includegraphics[width=\columnwidth]{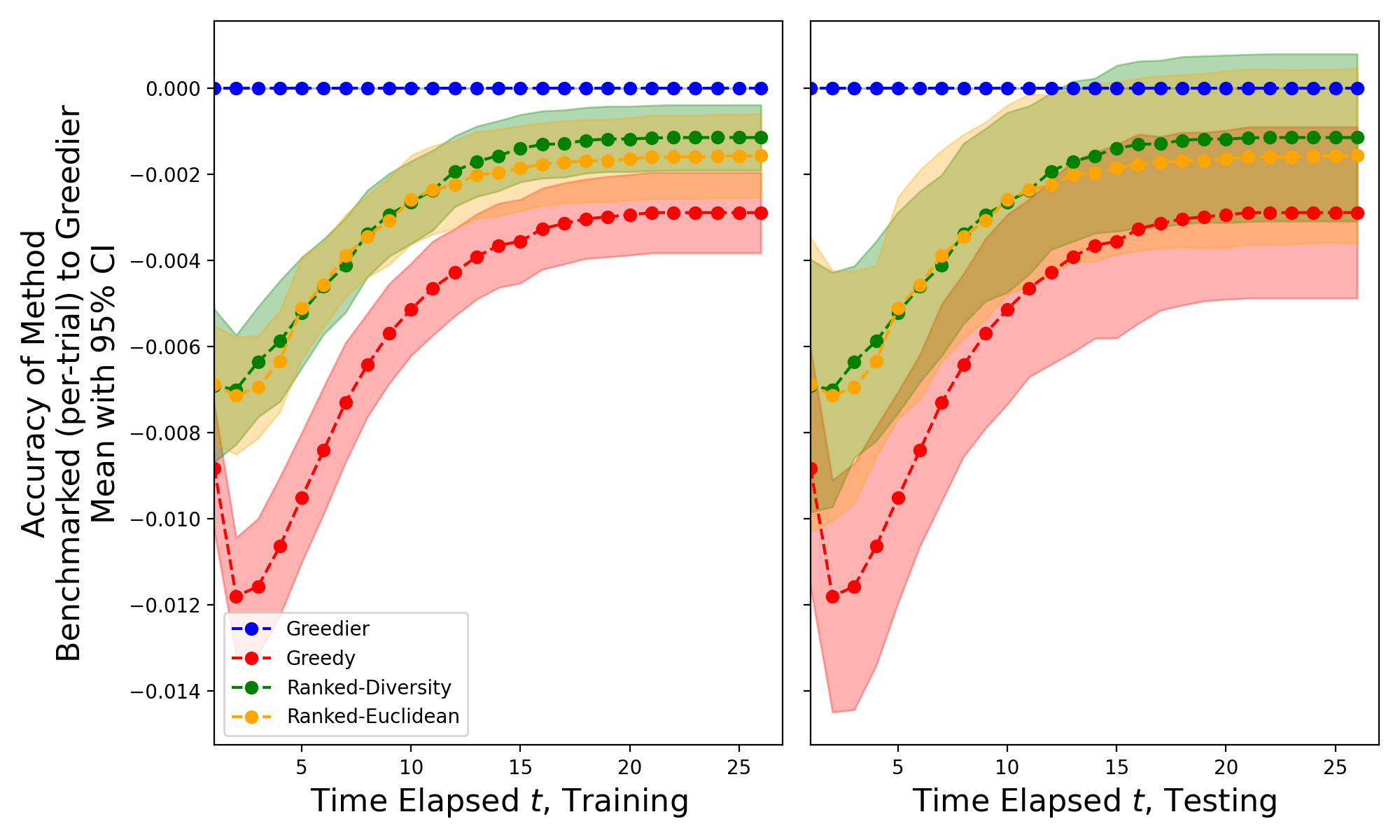}}
\vspace{-0.4cm}
\caption{Difference between greedier accuracy and other methods' accuracy averaged across all trials with 95\% confidence interval. Training at left, testing at right. Terminal value carried forward.}
\label{fig:accy}
\end{center}
\vskip -0.2in
\end{figure}

To compare performance the performance of the methods, for each trial in all test environments we calculate the difference in accuracy at each iteration from the other weight-ensembling algorithms to the greedier method, visualizing the average difference in Figure \ref{fig:accy} beginning when each WA has 2 ingredients. This amounts to benchmarking the performance of other methods relative to the greedier method. As some algorithms terminate before the ultimate time step $t=39$, we propagate the terminal accuracy value forward through time in order to still be able to run our aforementioned calculation for subsequent time steps. Initially, the increasingly negative values show that the greedier algorithm gains accuracy faster than the ranked and greedy methods at the outset. As more ingredients are included past $t=4$, ranked and greedy methods start to recover, with ranked methods suffering fewer losses and rebounding faster. This rebound occurs after many greedier runs have flatlined due to termination. Both ranked performances are similar, initially falling behind greedier but slowly recovering later on; greedy follows a similar trajectory, thus demonstrating that the greedier algorithm utilizes fewer ingredients effectively. ID validation (``training") accuracy for the non-greedier methods finish below the accuracy of greedier at a statistically significant level, while for OOD accuracy (``testing"), greedy closes below greedier at a statistically significant level with the ranked algorithms' intervals just barely enveloping the performance of greedier in the upper bound. 

\section{Explaining the Role of Diversity}

\subsection{Distributions of Quantiles over Selected Models}

\label{quantiles}

We next probe the extent to which distance measures (between the current WA to the candidate calculated ID when prediction is necessary) were associated with each algorithm's selection mechanism by binning the quantiles of distances of selected candidates to the current WA at each iteration time $t$. Since both greedier and ranked have a discrete host of models from which they can add a candidate at each step (but only select one), we first bin the quantile of the distance instead of the distance itself to make a well-posed statement about whether the more or less distant ingredients at some step were selected. For example, if $\theta_j$ had been the second least diverse from the current WA and had gotten selected for inclusion at time $t$, we would have binned $\frac{2}{{\textrm{Num Remaining}}}$ at time $t$. We can thus view each addition of a candidate as the discrete choice which was most useful to the WA.  A similar procedure may be repeated for the greedy algorithm, where we only advance $t$ in the figure when candidates are actually accepted by greedy; if a candidate is skipped (getting thrown out for the remainder of the procedure), we do not advance $t$ because in this visualization we are interested in the distances of \textit{selected} ingredients with the current WA. 

\begin{figure}[ht]
\vskip 0.2in
\begin{center}
\centerline{\includegraphics[width=1\columnwidth]{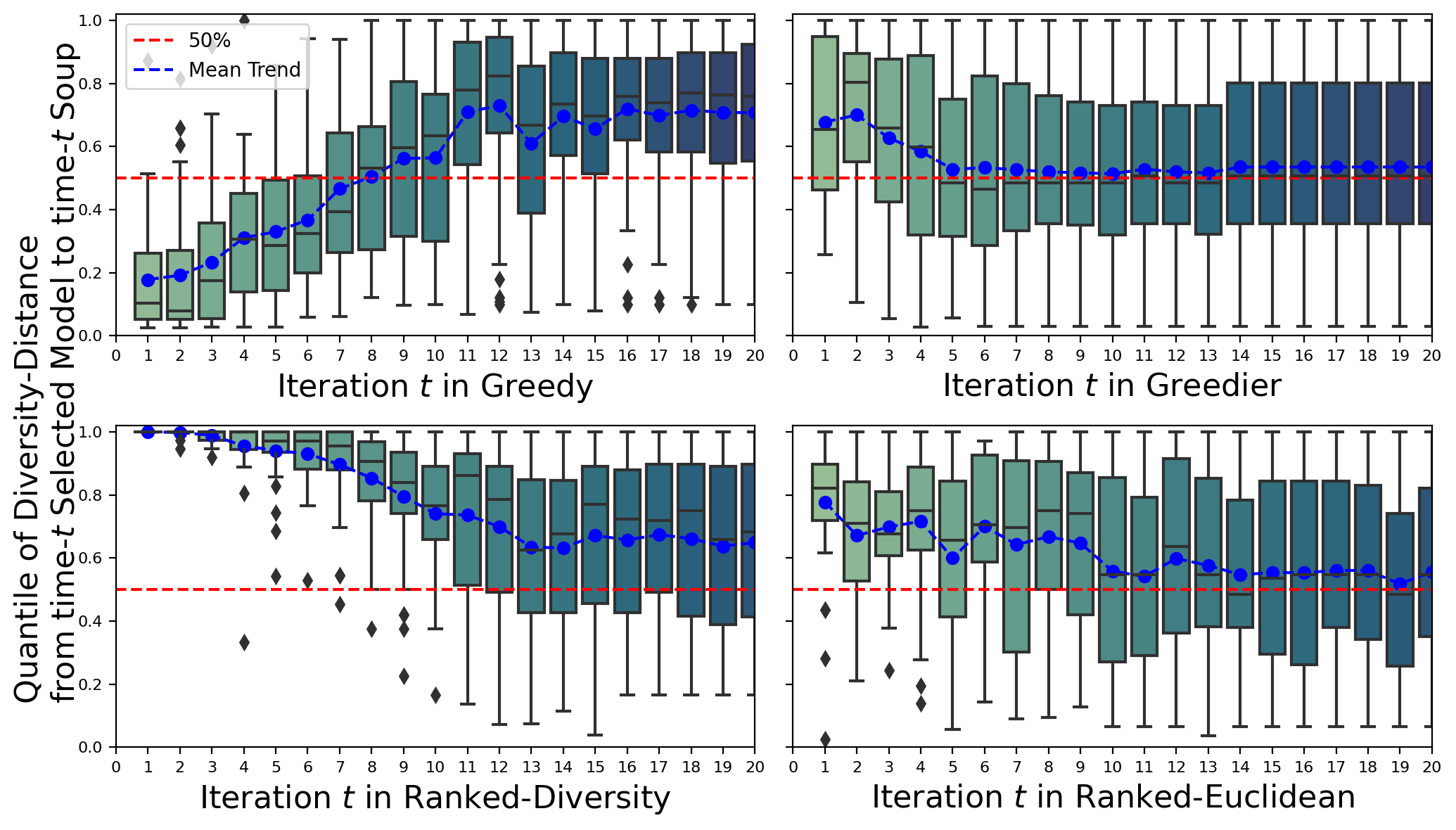}}
\caption{Box-plot of quantiles of diversity distance between the current WA and the selected model at each iteration $t$ of each algorithm across the 40 trials. Dashed red-line at 50\% indicates random selection.}
\label{fig:diversity-ranks}
\end{center}
\vskip -0.2in
\end{figure}

The distributions of the diversity-quantiles of selected models over time for each method is visualized in Figure \ref{fig:diversity-ranks}, with mean trends compared side-by-side in Figure \ref{fig:diversity-ranks-mean} of the Appendix. Over the greedier algorithm's first few iterations $t=1$ to $t=4$, we see that the distribution of quantiles is skewed to select ingredients which are \textit{more-diverse} than random chance, reinforced by the confidence interval in Figure \ref{fig:diversity-ranks-mean}. This comes in direct opposition to the greedy algorithm, which specifically selects \textit{less-diverse} candidates from its first iteration $t=1$ up to iteration $t=7$ as seen in Figure \ref{fig:diversity-ranks-mean}. As expected, the ranked-diversity algorithm tends to select ingredients which are most diverse from the current WA. 

Contextualizing early-stage diversity-quantile results with the ID and OOD accuracy gains that greedier and ranked make relative to greedy between $t=2$ to $t=4$ in Figure \ref{fig:accy}, this shows a clear association between selecting the most diverse points and rapid accuracy improvement. Yet, the algorithm designed to select the most diverse candidates, ranked-diversity, still underperforms the greedier method. This comes in spite of ranked-diversity having ingredient models which have the highest average pairwise diversity, the proxy of diversity used in \citet{rame2022diverse} as we see in Figure \ref{fig:avgpd-all}. Although ranked-diversity's selection of more diverse ingredients seen in Figure \ref{fig:diversity-ranks} led to ranked-diversity WAs having the highest average pairwise diversity as seen in Figure \ref{fig:avgpd-all}, the fact that the greedier method outperforms the ranked-diversity method while having a less diversity-inducing selection mechanism indicate that the greedier method benefits from some force beyond what is provided for by diversity. Diversity correlates with the benefits realized by the greedier algorithm, but it does not quite encapsulate the full power of the greedier algorithm because building it into the selection mechanism with ranked-diversity does not achieve competitive accuracy. That in the latter stages past $t=5$ of the greedier routine, both the diversity-quantiles of ingredients that we select for and the average pairwise diversity of the WA that is accepted seems to have saturated further evidence that the benefits of diversity are capped.

The quantiles of selected Euclidean distances are given by Figure \ref{fig:euclidean-ranks} in the Appendix. In the figure, we observe an even stronger association between high-quantile Euclidean distance and selection by our greedier algorithm, with each boxplot living well-above random chance up to $t=4$ when greedier's is making gains on algorithms' accuracies. The result is correlated with diversity selection, although clear differences in ranked-diversity and ranked-Euclidean distributions in Figures \ref{fig:diversity-ranks-mean}, \ref{fig:euclidean-ranks} indicate some decoupling between the selecting for divesity and Euclidean distance. The strong association of Euclidean distance and greedier decision-making demonstrate that sampling far-apart ingredients in a loss basin can be just as powerful as selecting for diversity when it is quantified by ratio-error.


\subsection{Dynamics of Errors}

In Section \ref{quantiles}, we concluded that diverse ingredients are a beneficial component for a WA, although WAs may improve through other means and the benefit stemming from diversity is likely to saturate relatively quickly. In this section, we analyze the dynamics of how the errors that WAs make evolve as ingredients are added to the WA. The comparison of such dynamics across algorithms will make conspicuous the direct benefits and limitations of diversity for ID and OOD prediction, and how the greedier approach may exploit new ingredients better than ranked-diversity.

To this end, at each time step $t$ (starting with $t=1$ with 2 ingredients) we split up the ID and OOD sets into data points disjointly into four sets: 1) points which the WA at time $t$ had classified incorrectly but the time-$t$ selected ingredient classified correctly (not yet included in the latter WA), denoted ``$t$-incorrect ingredient-correct", and similarly the disjoint sets 2) ``$t$-correct ingredient-incorrect", 3) ``$t$-correct ingredient-correct", and 4) ``$t$-incorrect and ingredient-correct". For each of these sets are interested in the probability for some data point that the ingredient's outcome takes hold in the $t+1$ WA: such as for 1) the probability that the time $t+1$ WA is correct given that the time $t$ WA was incorrect and the ingredient was correct. 

\begin{figure}[ht]
\vskip 0.2in
\begin{center}
\centerline{\includegraphics[width=1\columnwidth]{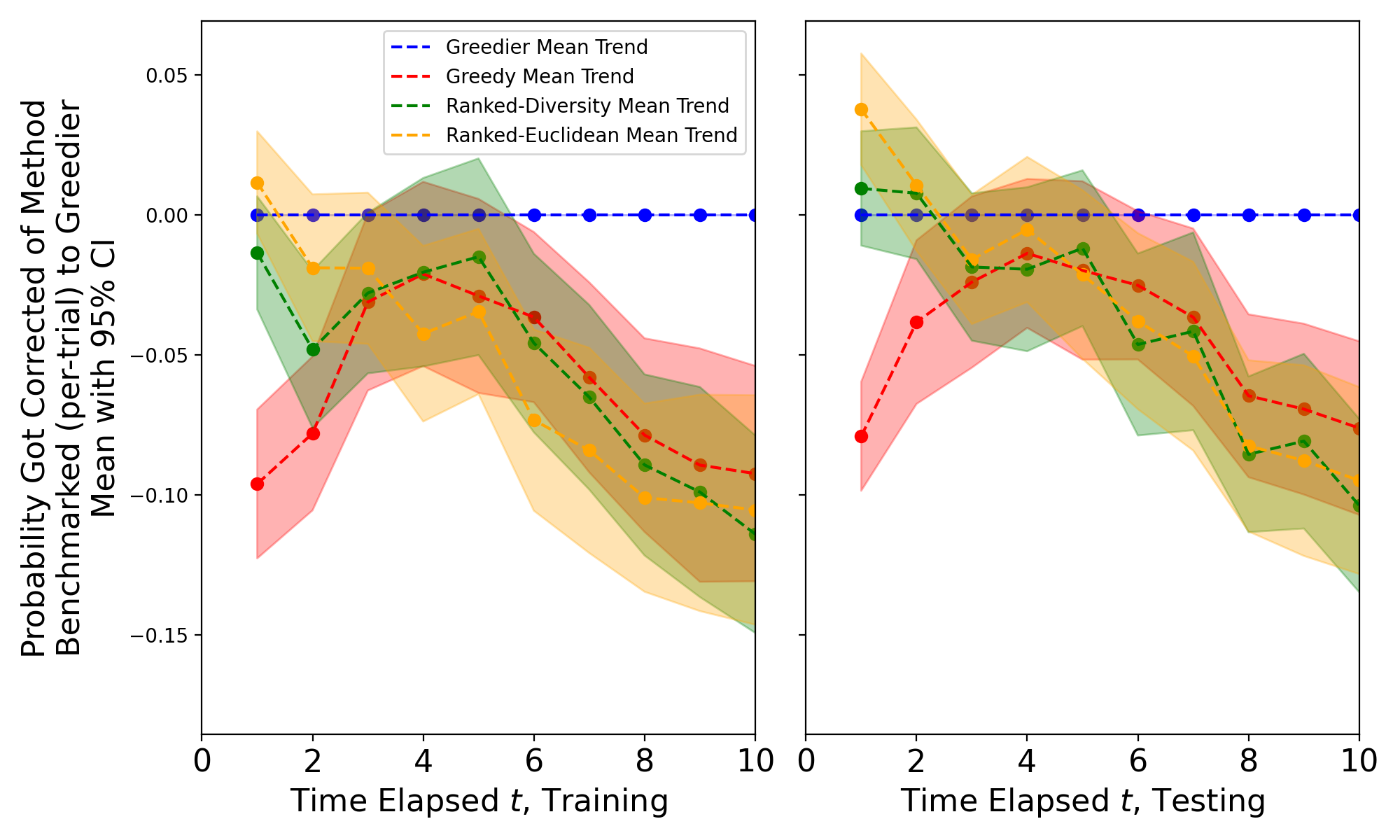}}
\caption{$t$-incorrect ingredient-correct: Probabilities that the next-step WA predicts correctly given that the current-step WA was incorrect and the ingredient was correct, difference from greedier and other methods' averaged across all trials with 95\% confidence interval. Training at left, testing at right. Terminal value carried forward.}
\label{fig:got-corrected-10}
\end{center}
\vskip -0.2in
\end{figure}

As in previous analysis, given a weight-ensembling method and experimental trial, we benchmark each series with respect to the greedier method's series to contrast the methods. We carry forward the terminal values of each series before averaging. In Figure \ref{fig:got-corrected-10}, we plot the first 10 iterations of the differenced series corresponding to $t$-incorrect ingredient-correct. We plot the first ten only as the greedier method typically terminates within 10 time steps. For these early time periods, it is clear that the greedier algorithm makes relatively better use of its new ingredients in both the ID and OOD setting. Noticeably, we observe the importance of diversity for generalization in the OOD setting (right), where for the first several iterations algorithms ($t=2, t=3$ when the impact of adding a new ingredient is greatest due to weighting), selecting for diversity induces approximately a 10\% greater probability of being the current-step WA's mistake corrected by the ingredient. The decreasing trend of other non-greedier beyond $t=5$ correspond to many cases in which greedier has terminated, and other methods (which may still be running) already include more ingredients and thus it is more difficult for any newly-added individual to strongly impact the result. Figures \ref{fig:got-incorrected-10}, \ref{fig:remain-incorrect-10} in the appendix demonstrate that the selection of diverse candidates makes scenarios 2) and 4), in which new errors appear or existing ones are retained, less likely. 


\section{Visualization of Weight-Ensembling}

\subsection{MDS Visualization of the Greedier Algorithm}

We develop a visualization method to capture all pairs of distance relationships jointly without reducing the diversity to a single number, as does average pairwise diversity. We do so with Multidimensional Scaling (MDS), a dimensionality reduction algorithm designed to preserve pairwise distances.

After running a weight-ensembling algorithm on the set of $k$ neural networks, we calculate all pairwise distances between the models evaluated in the experiment. We store these distances in a symmetric distance matrix, then use this matrix as a plug-in to MDS. We use metric MDS for Euclidean distance, and for diversity we use non-metric MDS. At each iteration $t$, we reveal in the decomposed space the candidate WAs that we considered at this time using our current set of ingredients and the remaining ingredients. Due to the structure of MDS, we thus visualize the pairwise distances between the candidate WAs, the current and past WAs, and the individual candidates which shed qualitative insight on the selection mechanism of the greedier algorithm. 

We demonstrate the progression of the greedier algorithm through Euclidean distance in one example in Figure \ref{fig:euclidean-MDS-stacked} (in Appendix due to figure size).  Consistent with the results of quantile-ranks in Euclidean space, we see for iterations $t = 2, 3, 4$ that we have selected points in space which were on the larger end of Euclidean distance from the current WA; at $t=5$ we have saturated accuracy and the algorithm has terminated. We can see convergence in decomposed weight space, reflecting the diminishing returns from adding more candidates after location in Euclidean space saturates. Turning to the diversity distance Figure \ref{fig:diversity-MDS-stacked} we observe that the WA points do not converge as nicely as time passes through the algorithm, although we still manage to select candidates which tend to lie on the farther side from our current WA through 4 iterations. Finally, we observe the intuition allotted by the visualization technique: we have selected truly distinct points in weight space, as the selected candidates exhibit separation in both decomposed spaces, as opposed to the comparitively reductive average pairwise diversity or Euclidean distance in the literature.

\vspace{-0.3cm}

\section{Discussion}

We have introduced the greedier algorithm which at each iteration selects the ingredient which maximally improved the WA's ID validation performance. When treated as a ``gold-standard," greedier's decision-making helps to uncover how relationships between ingredients and a WA are leveraged to best improve the WA's performance. We also propose the ranked algorithm, which at each iteration sorts ingredients by their distance from the WA and selects the first to improve ID validation accuracy. We can contrast greedier results with the ranked and greedy algorithms, using different behaviors to reason about the role that diversity plays in performance dynamics. Leveraging this structure, we identify that both high diversity distance and high Euclidean distance explain the selection method when performance improves the fastest, implying that selecting diverse or spaced out candidates contributes rapidly towards improving the WA. Yet, that the ranked-diversity algorithm does not match the greedier method ID or OOD limit the extent to which diversity plays a role in this performance. We finally introduce a method by which we can examine how our algorithms selection traverses a loss basin and whether our candidates are truly diverse by not reducing our distance relationships but rather leveraging pairwise structure in a decomposition for qualitative analysis.

\section*{Impact Statement}

This work provides a new weight-ensembling algorithm and explanations for why a WA procedure improves accuracy when able to choose which model to add to the list of candidates using the greedier algorithm. By shedding light on what may cause WA to improve, we provide new support to an existing avenue by which they can improve inference using the WA efficiently. Such work has the potential to improve deep learning algorithms in all facets of society, both for clearly positive (such as medicine) and more nuanced to negative (such as surveillance) purposes.





\bibliography{main}
\bibliographystyle{icml2024}

\newpage
\appendix
\onecolumn





\section{Quantiles and Distance Distributions from Iterations of Weight-Ensemble Algorithms to Selected Ingredients}

In this section, we provide the boxplots of distances and quantiles of the selected models at each iteration using our weight-space geometric distance measures. We also plot their mean trends together for a closer comparison.

\subsection{Distance Distributions}

In Section \ref{quantiles}, the quantiles of the distances of selected ingredients from a current soup has the advantage of elucidating within a set of ingredients whether the more or less diverse ingredients were utile. In contrast, we examine here the raw trends (without taking quantiles) to ensure the robustness of the analysis.

For example, we can explain the first iteration of this process through the lens of diversity distance. Following notation in Algorithm \ref{alg:greedier}, at $t=1$ we know the current WA is equal to the model $\textrm{average}(\textrm{ingredients}) = \theta_1$. Then if we selected $\theta_j$ to add to the ingredients using the greedier method, we store the result $d_D(\theta_1, \theta_j)$ in our bin for $t=1$. We proceed this binning from $t=1$ to $t=T_{max}$ where $T_{max}$ is equal to the largest amount of time any greedier algorithm instance ran for. We then boxplot over each $t$. As in the quantile example, we have an analogous implementation for the greedy algorithm. These results are similarly in favor of the higher diversity and Euclidean distances being selected by the greedier algorithm in earlier iterations. In both but especially in the Euclidean res ult of Figure \ref{fig:euclidean-raw} we see a steep drop-off in selected distances after $t = 4$. Such a dropoff is intuitive because as we roughly move towards center of the the points in weight space, our distance to unincorporated but likely related points will also decrease. 

\subsection{Diversity Distance}

\begin{figure}[ht]
\vskip 0.2in
\begin{center}
\centerline{\includegraphics[width=\columnwidth]{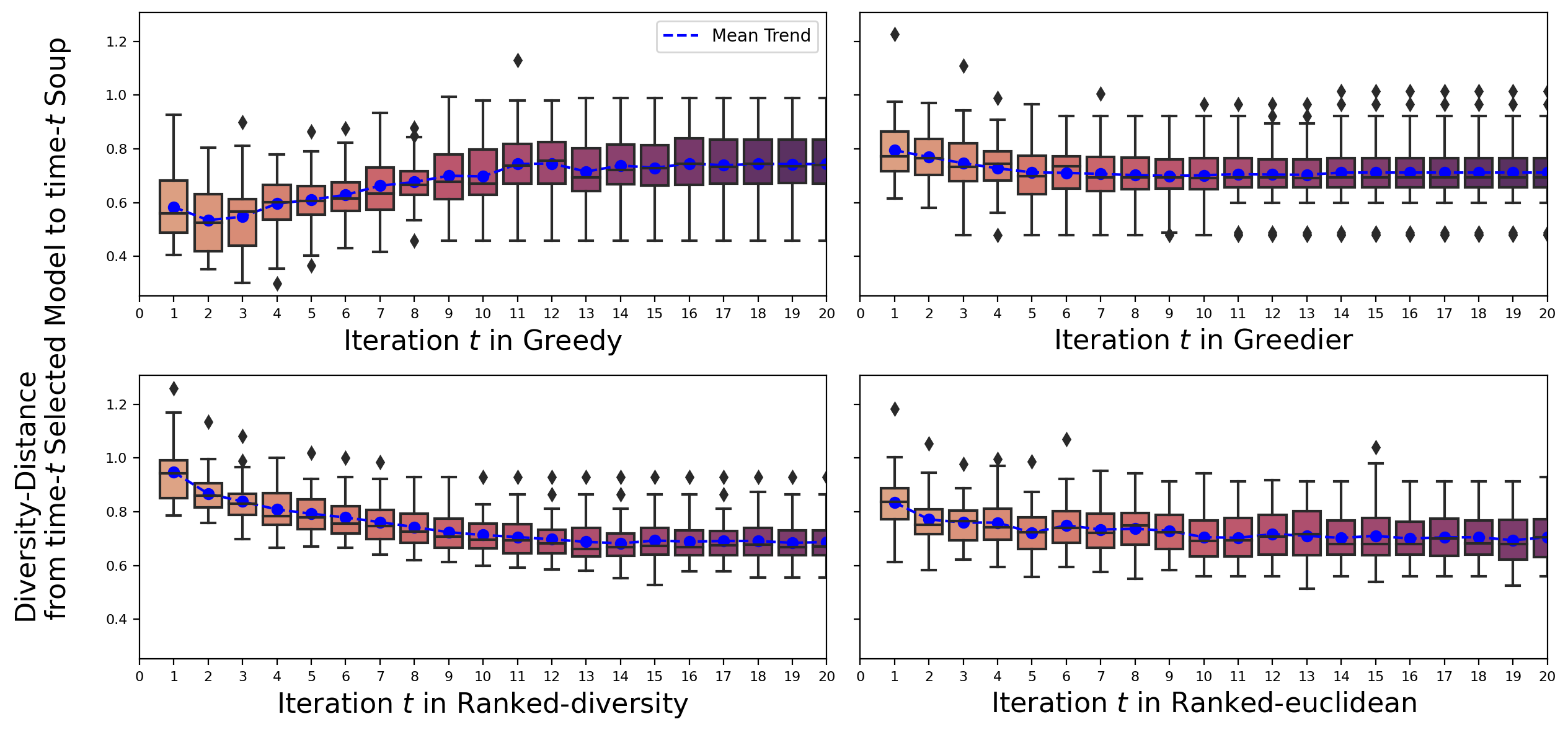}}
\caption{Box-plot of diversity distance between the current WA and the selected model at each iteration $t$ of each algorithm across the 40 trials. }
\label{fig:diversity-raw}
\end{center}
\vskip -0.2in
\end{figure}

\begin{figure}[ht]
\vskip 0.2in
\begin{center}
\centerline{\includegraphics[width=.75\columnwidth]{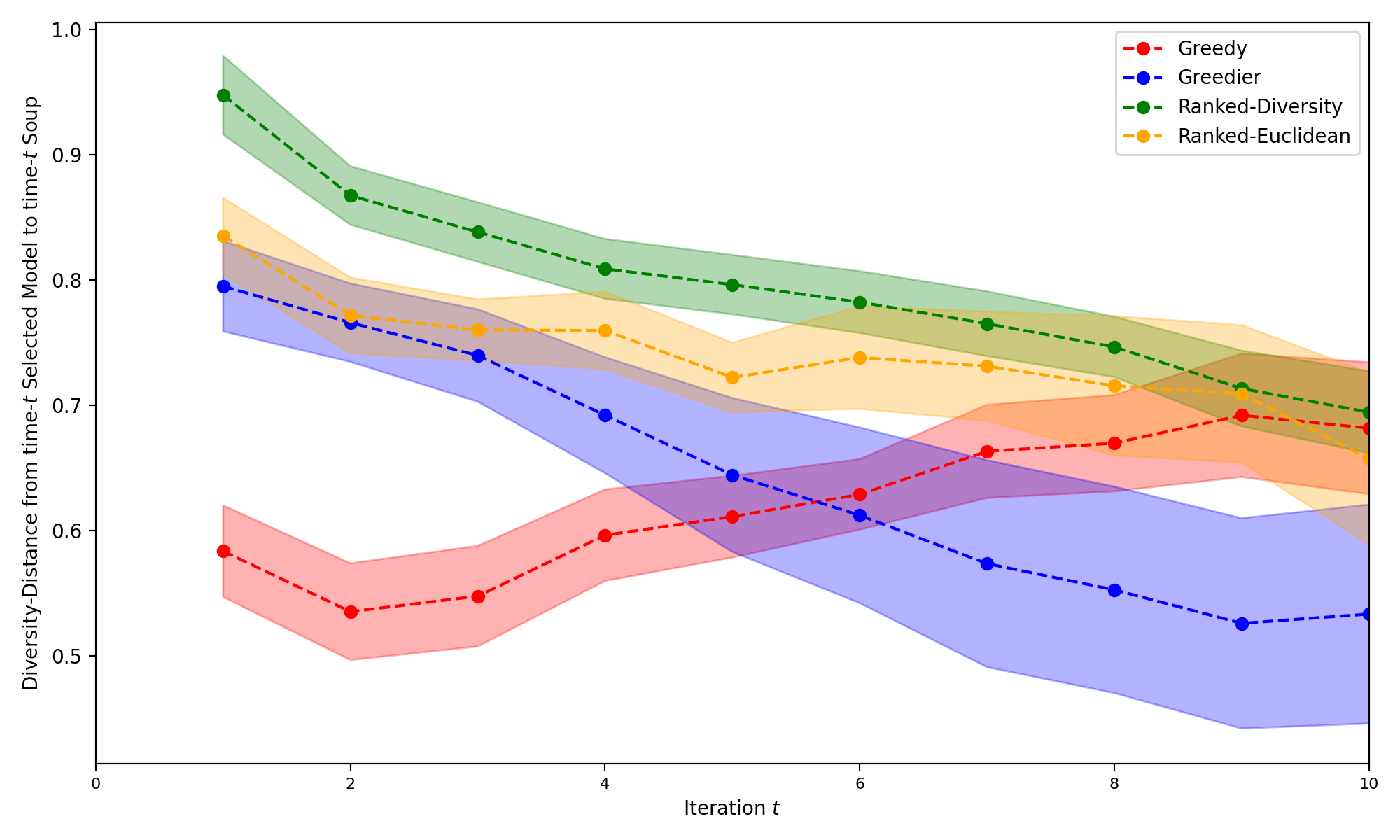}}
\caption{Diversity distance between the current WA and the selected model at each iteration $t$ of the greedy, greedier, ranked-diveristy, and ranked-Euclidean algorithms averaged across all trials with 95\% confidence interval. First ten iterations plotted.}
\label{fig:diversity-raw-mean}
\end{center}
\vskip -0.2in
\end{figure}

\begin{figure}[ht]
\vskip 0.2in
\begin{center}
\centerline{\includegraphics[width=.75\columnwidth]{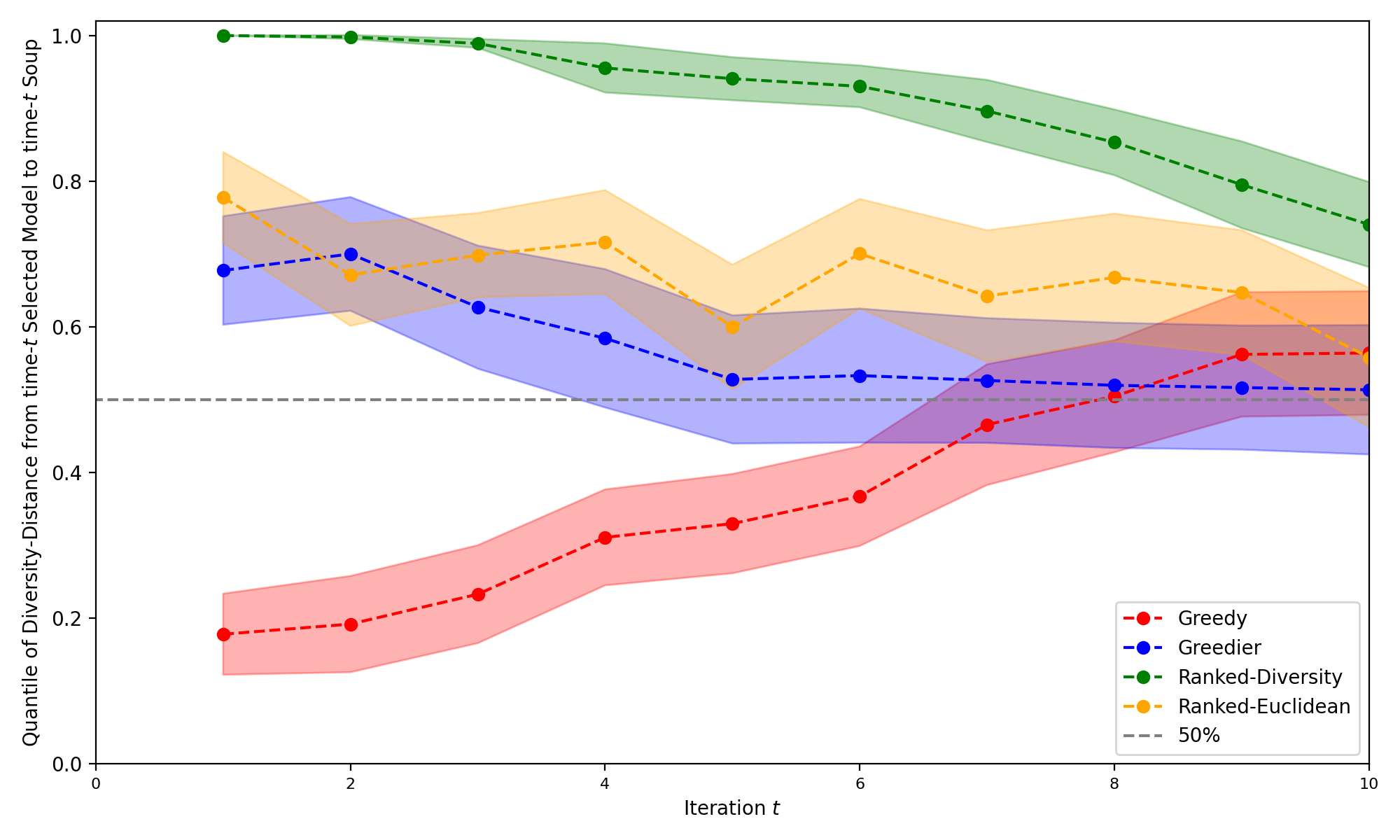}}
\caption{Quantiles of Diversity distance between the current WA and the selected model at each iteration $t$ of the greedy, greedier, ranked-diveristy, and ranked-Euclidean algorithms averaged across the 40 runs with 95\% confidence interval. First ten time steps plotted.}
\label{fig:diversity-ranks-mean}
\end{center}
\vskip -0.2in
\end{figure}

The distributions over selected-ingredient diversity distances is visualized in Figure \ref{fig:diversity-raw}. We also plot the mean trends in selected diversity distance with confidence intervals in one plot in Figure \ref{fig:diversity-raw-mean}. Elaborating on Figure \ref{fig:diversity-ranks}, we plot the mean trend in the quantiles of the distances of selected ingredients in one plot in Figure \ref{fig:diversity-ranks-mean}.

\subsection{Euclidean Distance}

\begin{figure}[ht] 
\vskip 0.2in
\begin{center}
\centerline{\includegraphics[width=.8\columnwidth]{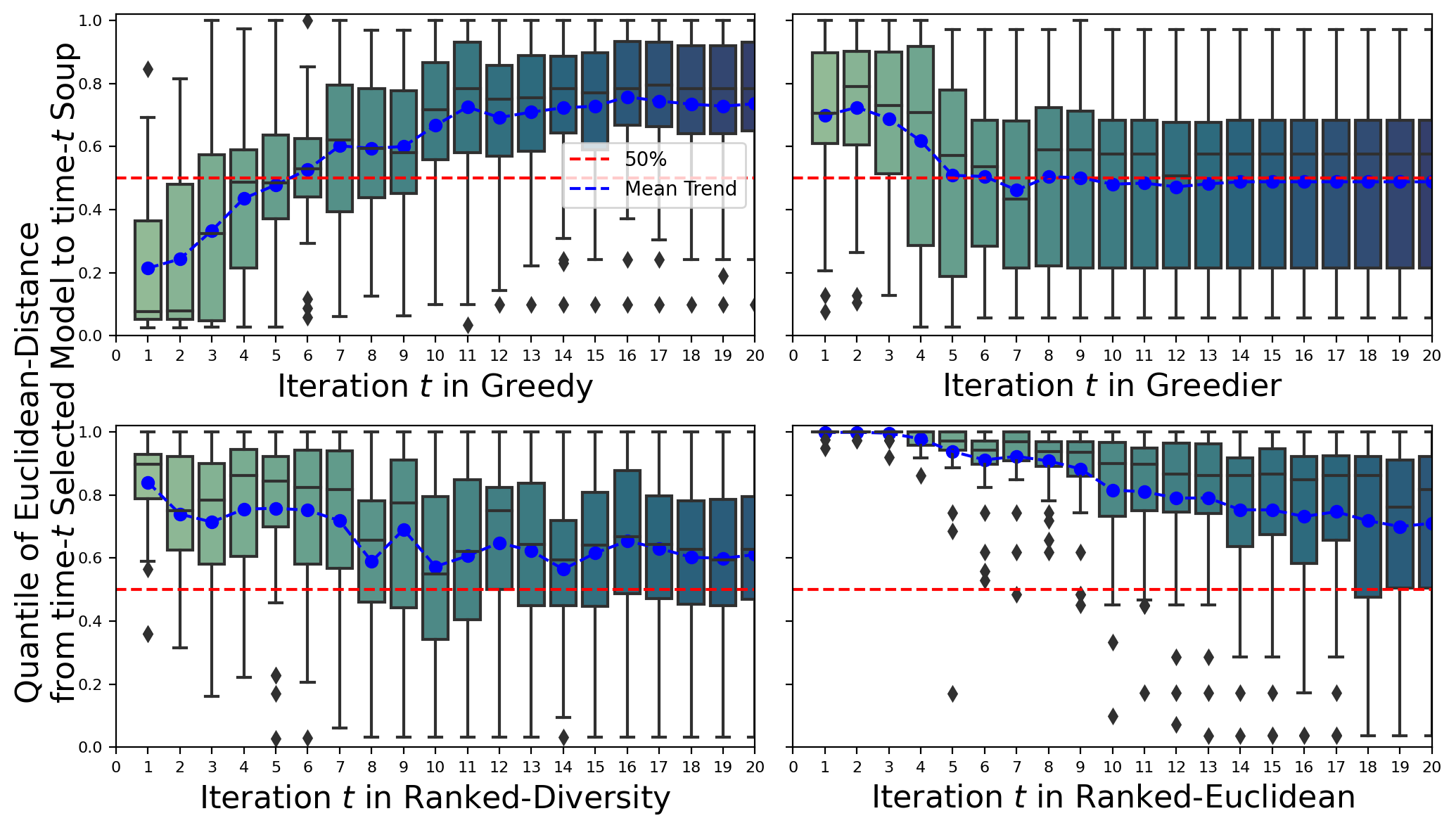}}
\caption{Box-plot of quantiles of Euclidean distance between the current WA and the selected model at each iteration $t$ of each algorithm across the 40 trials. Dashed red-line at 50\% indicates random selection.}
\label{fig:euclidean-ranks}
\end{center}
\vskip -0.2in
\end{figure}

\begin{figure}[ht] 
\vskip 0.2in
\begin{center}
\centerline{\includegraphics[width=\columnwidth]{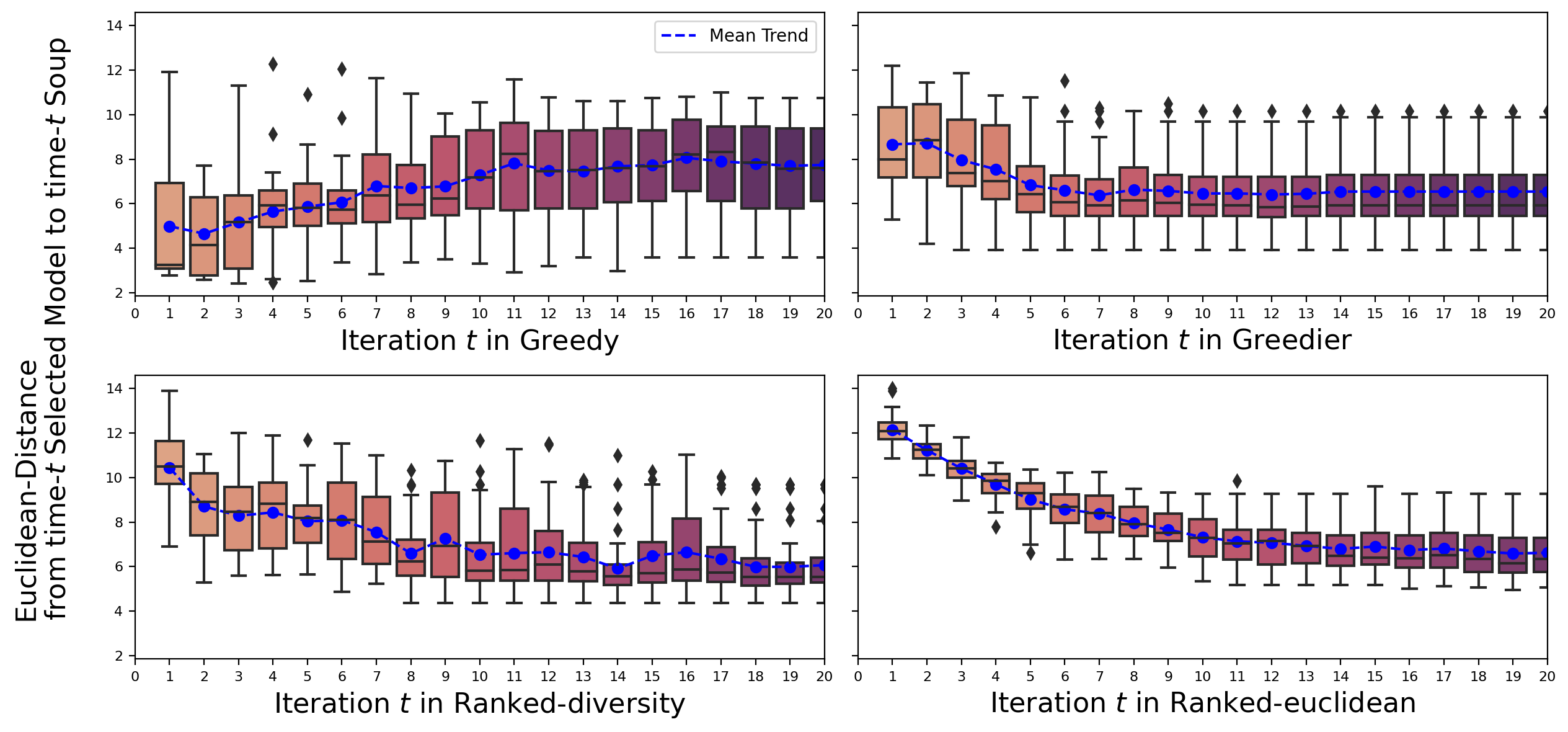}}
\caption{Box-plot of Euclidean distance between the current WA and the selected model at each iteration $t$ of each algorithm across the 40 trials.}
\label{fig:euclidean-raw}
\end{center}
\vskip -0.2in
\end{figure}

\begin{figure}[ht]
\vskip 0.2in
\begin{center}
\centerline{\includegraphics[width=.75\columnwidth]{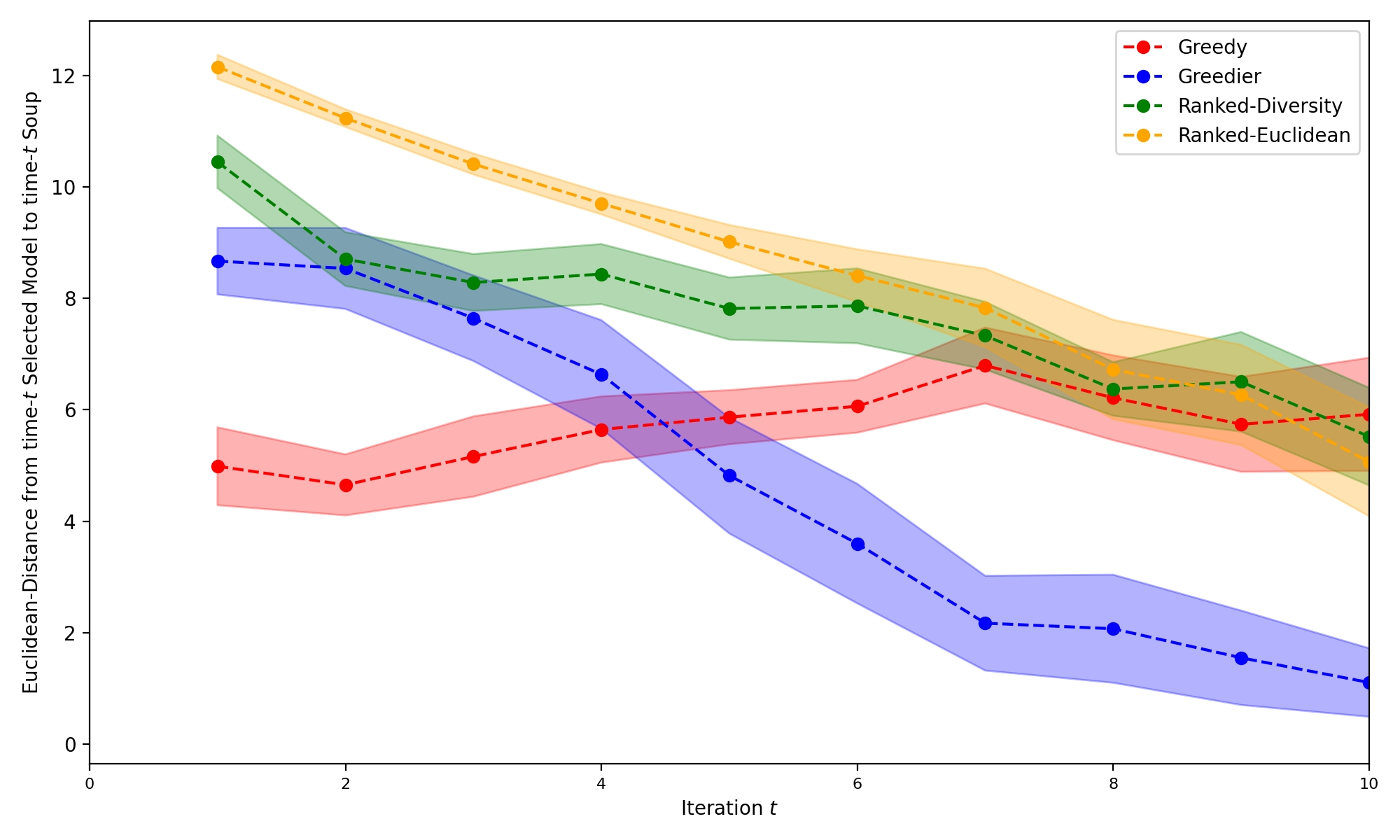}}
\caption{Euclidean distance between the current WA and the selected model at each iteration $t$ of the greedy, greedier, ranked-diveristy, and ranked-Euclidean algorithms averaged across all trials with 95\% confidence interval. First ten iterations plotted.}
\label{fig:euclidean-raw-mean}
\end{center}
\vskip -0.2in
\end{figure}

\begin{figure}[ht]
\vskip 0.2in
\begin{center}
\centerline{\includegraphics[width=.75\columnwidth]{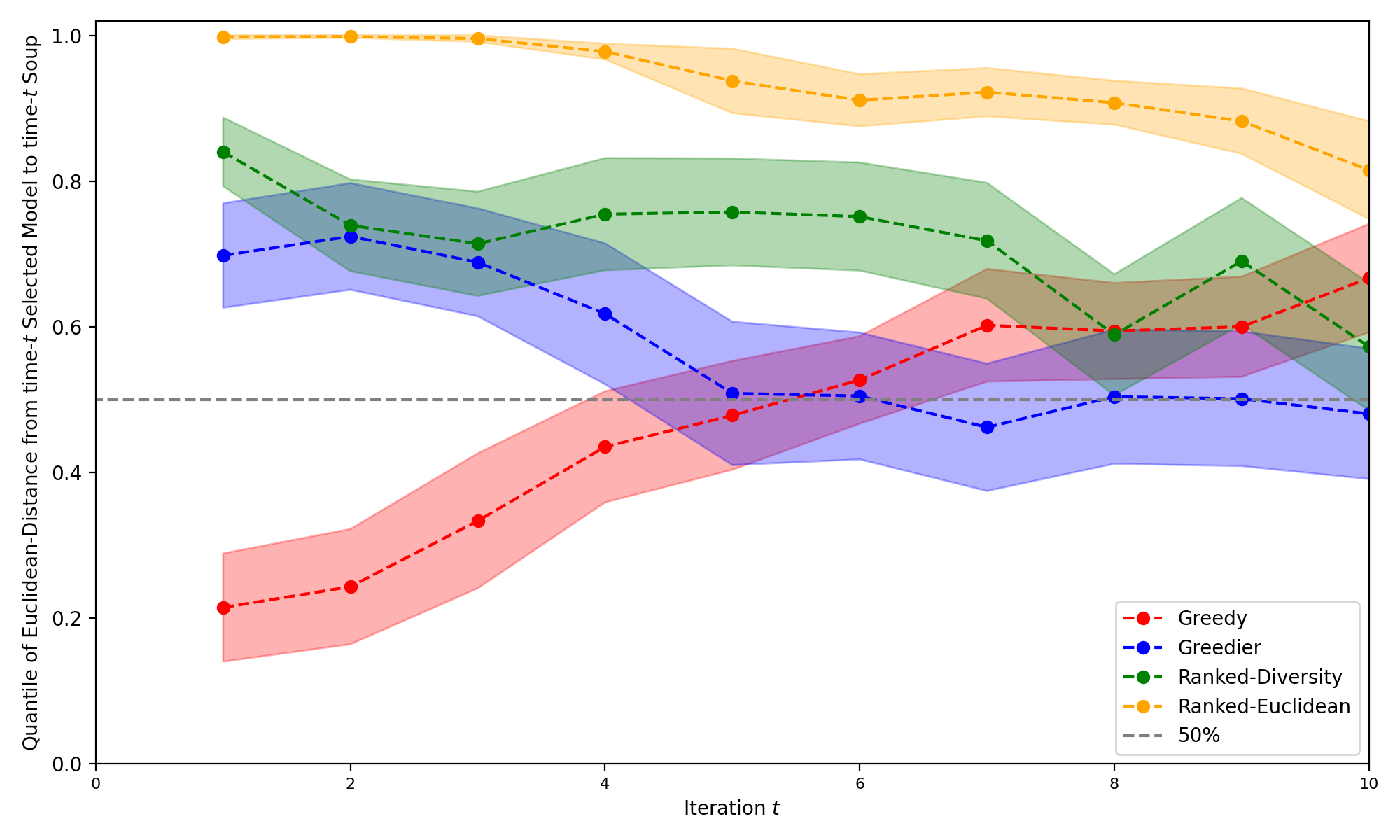}}
\caption{Quantiles of Euclidean distance between the current WA and the selected model at each iteration $t$ of the greedy, greedier, ranked-diveristy, and ranked-Euclidean algorithms averaged across the 40 runs with 95\% confidence interval. First ten time steps plotted.}
\label{fig:euclidean-ranks-mean}
\end{center}
\vskip -0.2in
\end{figure}

The quantiles over selected-ingredient Euclidean distances is visualized in Figure \ref{fig:euclidean-ranks}. The distributions over selected-ingredient Euclidean distances is visualized in Figure \ref{fig:euclidean-raw}. We also plot the mean trends in selected diversity distance with confidence intervals in one plot in Figure \ref{fig:euclidean-raw-mean}. Elaborating on Figure \ref{fig:euclidean-ranks}, we plot the mean trend in the quantiles of the distances of selected ingredients in one plot in Figure \ref{fig:euclidean-ranks-mean}.

\section{MDS Extended}

\begin{figure}[ht] 
    \centering
    \begin{subfigure}
        \centering
        \includegraphics[width=.88\columnwidth]{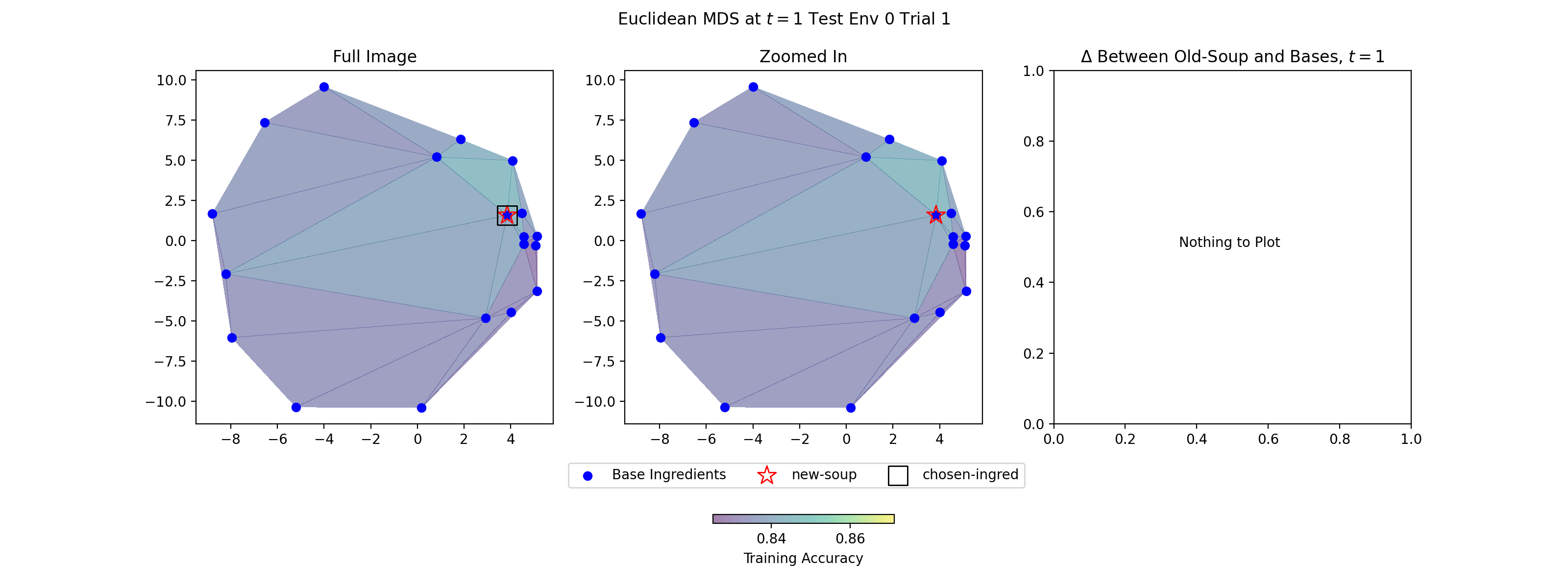}
    \end{subfigure}
    \vskip -0.3in
    \begin{subfigure}
        \centering
        \includegraphics[width=.88\columnwidth]{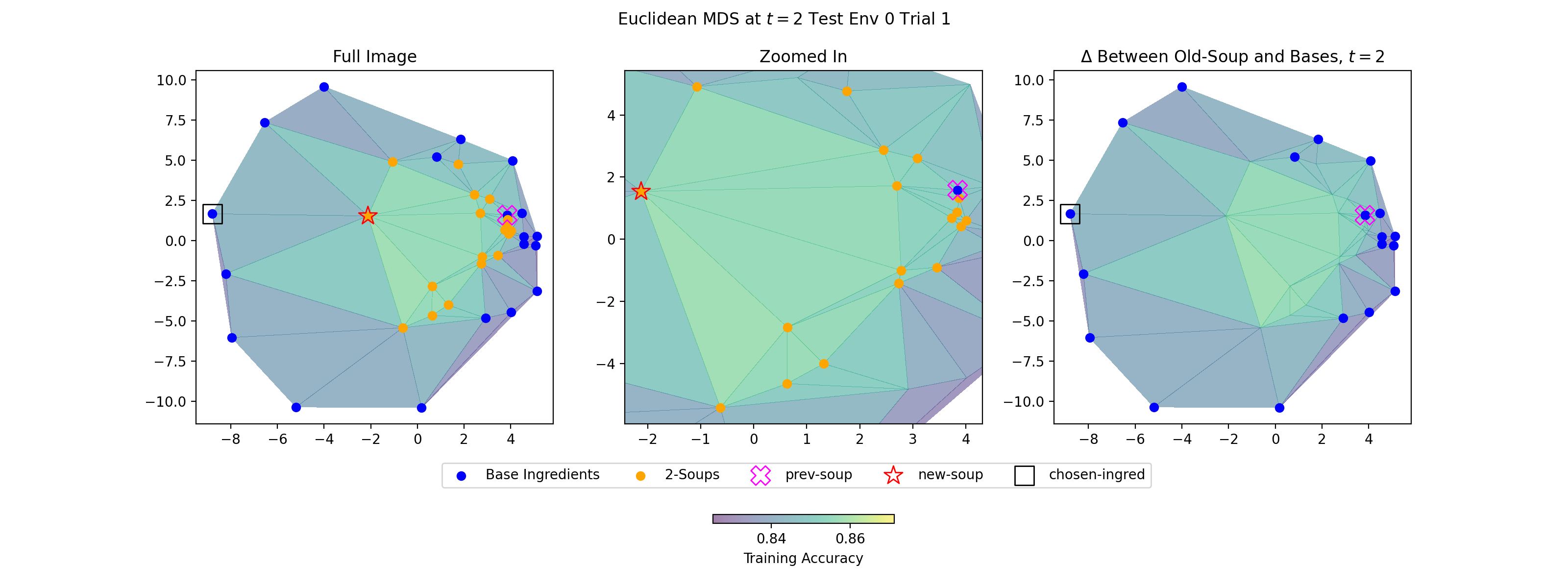}
    \end{subfigure}
    \vskip -0.3in
    \begin{subfigure}
        \centering
        \includegraphics[width=.88\columnwidth]{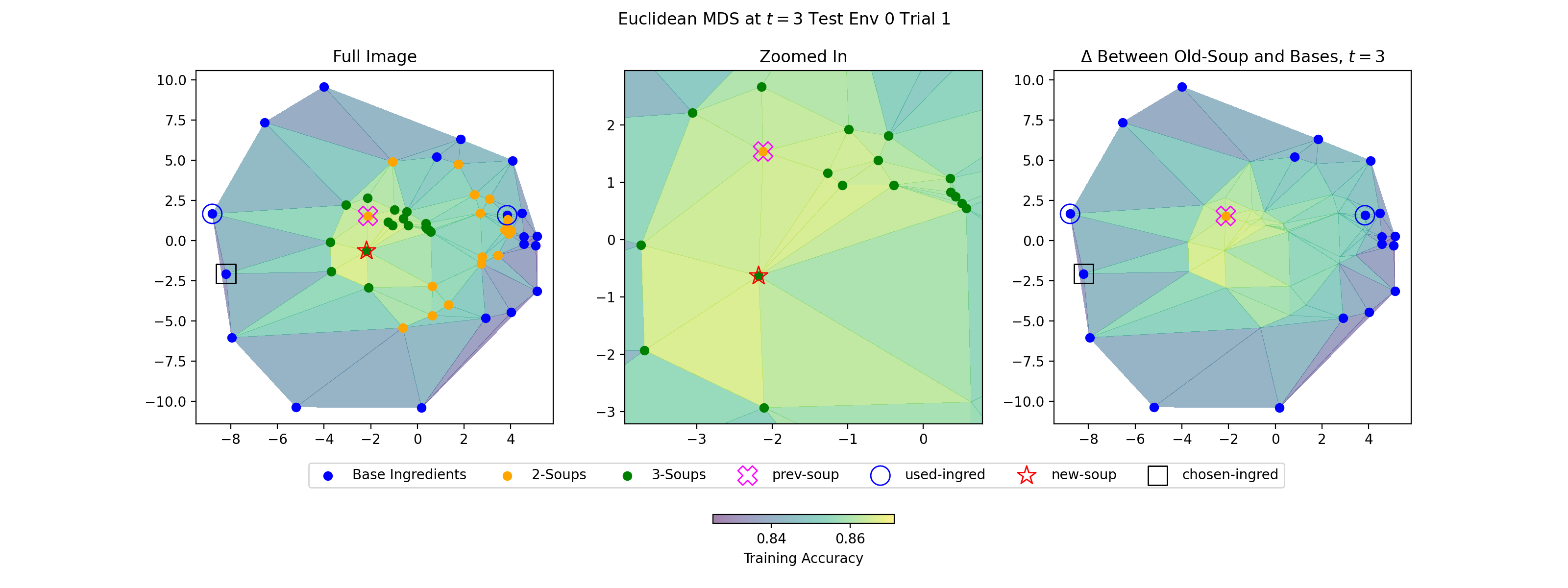}
    \end{subfigure}
    \vskip -0.3in
    \begin{subfigure}
        \centering
        \includegraphics[width=.88\columnwidth]{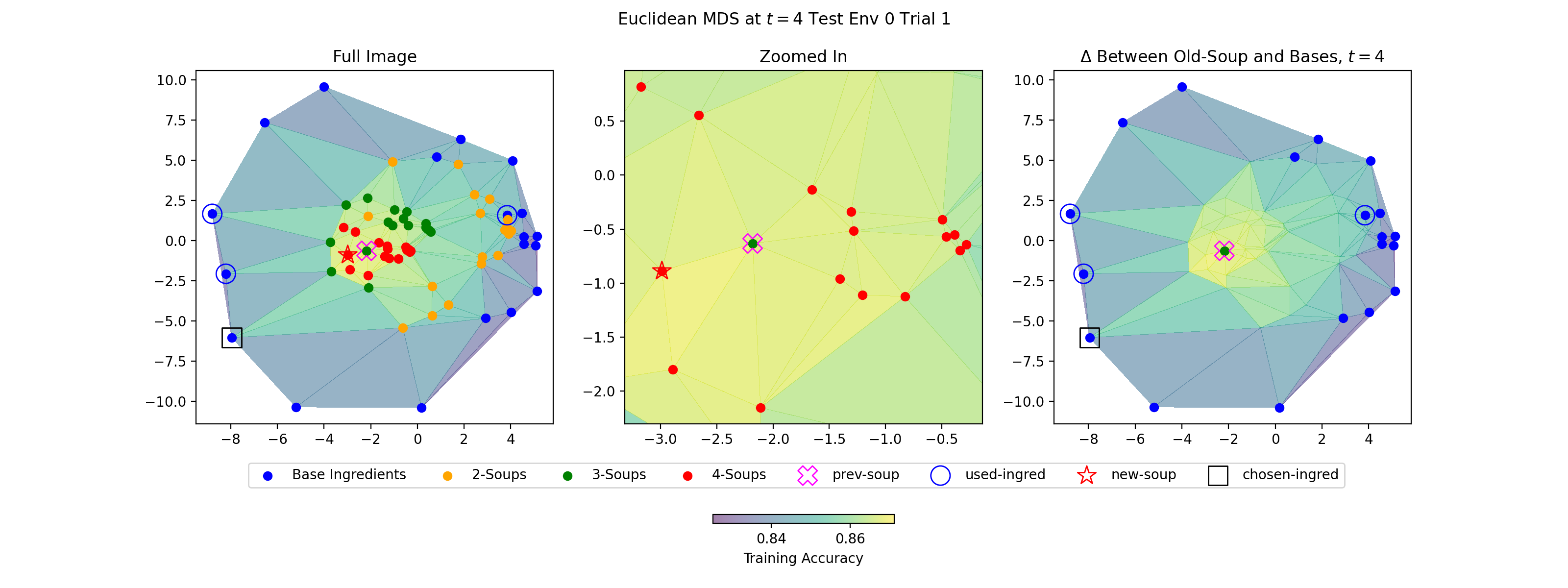}
    \end{subfigure}
    
    \caption{Euclidean distances between models in the greedier procedure plugged into MDS, with points color-coded by number of ingredients. Previously used ingredients are circled, currently selected ones are in a square, and the current WA is in an x. Backdropped by triangulated accuracy. Left: all points up to and including time $t$. Center: zoomed in on previous WA and time $t$ candidate WAs. Right: current WA and individual candidates. Smaller experiment (20 candidates) is visualized.}
    \label{fig:euclidean-MDS-stacked}
\end{figure}

\begin{figure}[ht]
    \centering
    \begin{subfigure}
        \centering
        \includegraphics[width=.88\columnwidth]{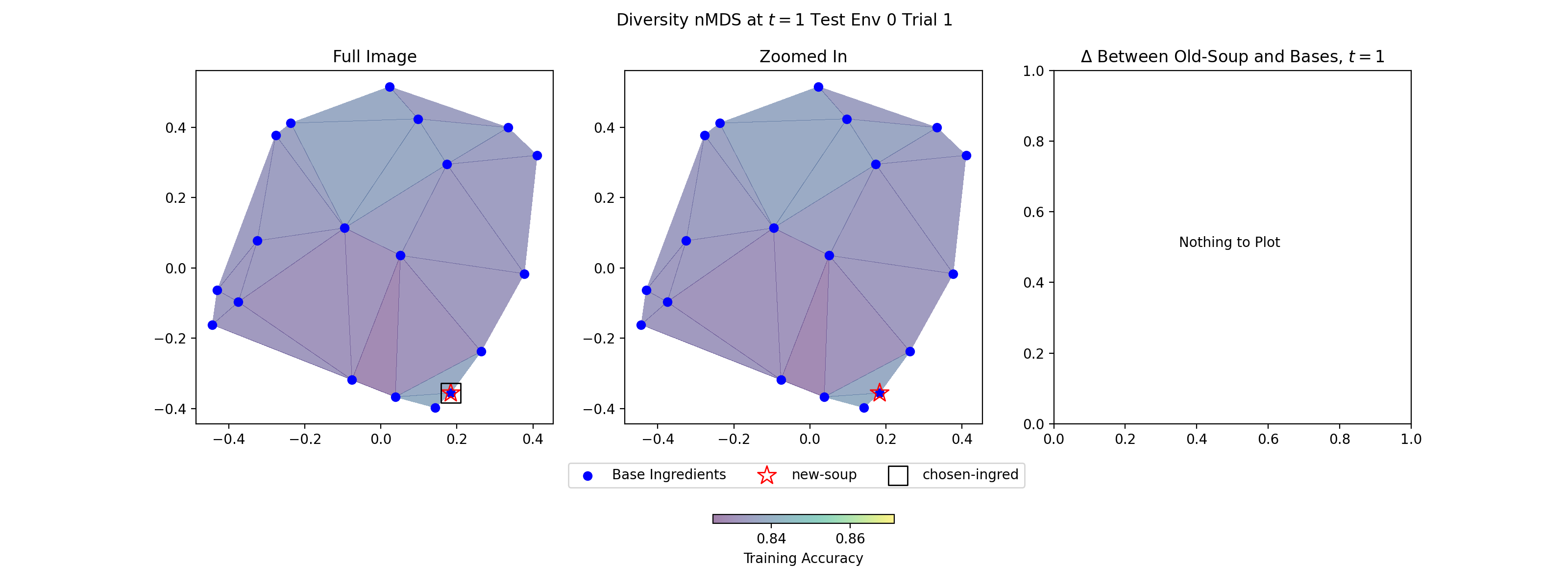}
    \end{subfigure}
    \vskip -0.3in
    \begin{subfigure}
        \centering
        \includegraphics[width=.88\columnwidth]{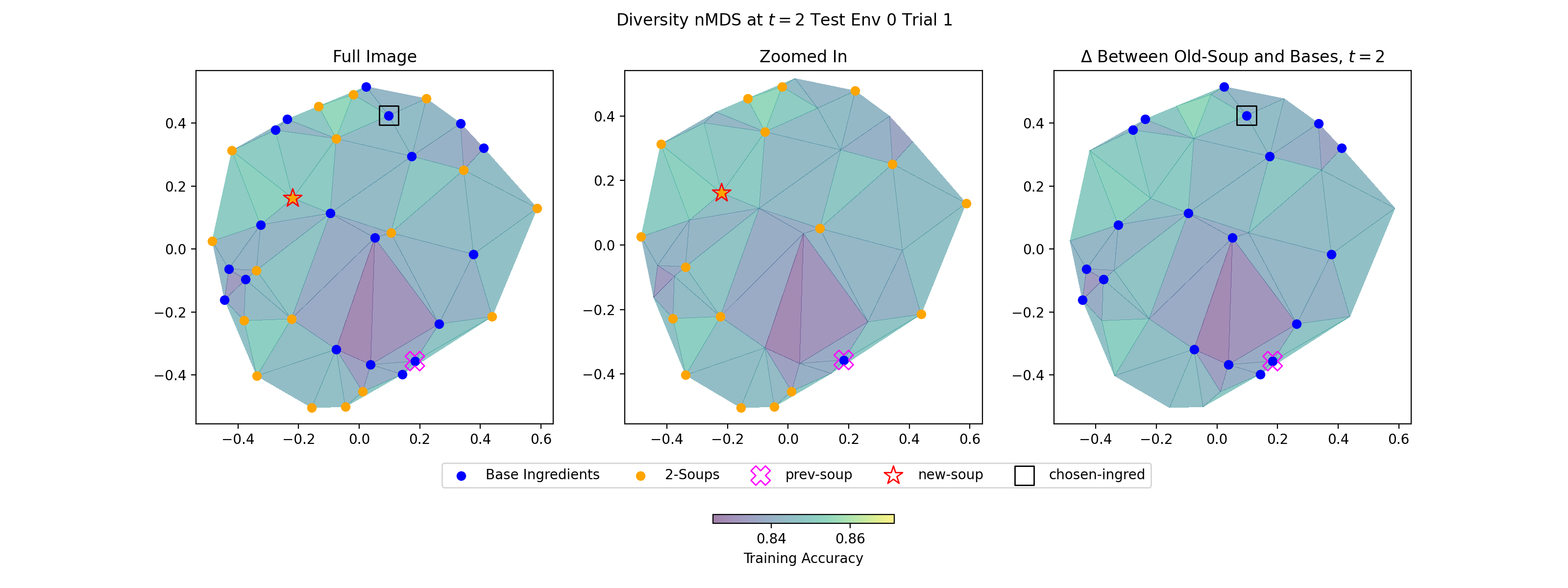}
    \end{subfigure}
    \vskip -0.3in
    \begin{subfigure}
        \centering
        \includegraphics[width=.88\columnwidth]{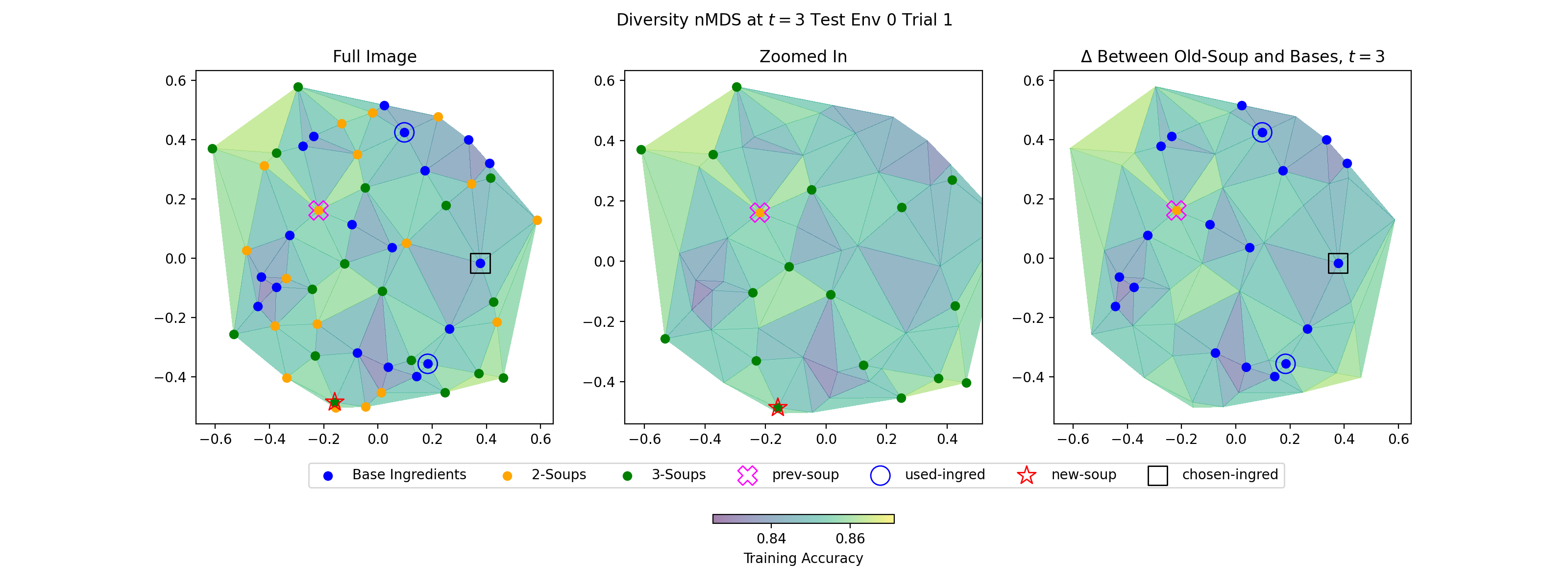}
    \end{subfigure}
    \vskip -0.3in
    \begin{subfigure}
        \centering
        \includegraphics[width=.88\columnwidth]{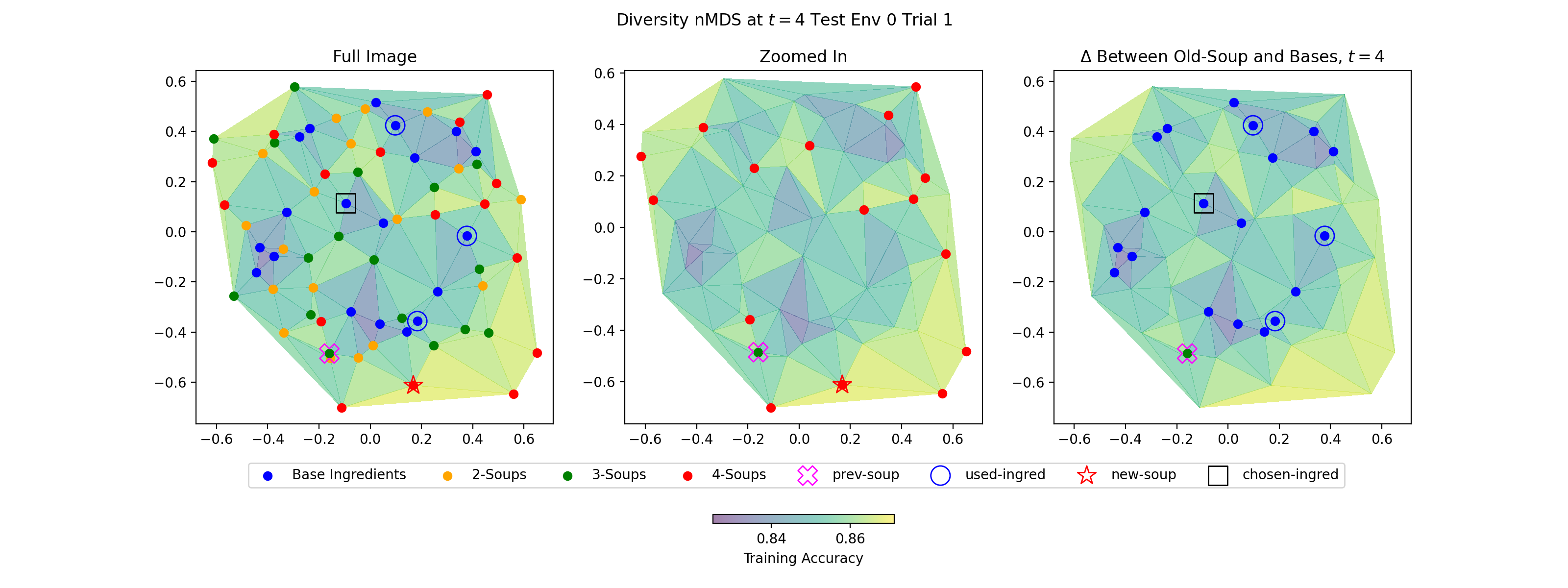}
    \end{subfigure}
    
    \caption{Diversity distances between models in the greedier procedure plugged into non-metric MDS, with points color-coded by number of ingredients. Previously used ingredients are circled, currently selected ones are in a square, and the current WA is in an x. Backdropped by accuracy. Left: all points up to and including time $t$. Center: zoomed in on previous WA and time $t$ candidate WAs. Right: current WA and individual candidates. Smaller experiment (20 candidates) is visualized.}
    \label{fig:diversity-MDS-stacked}
\end{figure}

We demonstrate an example of the MDS visualization in Euclidean space (Figure \ref{fig:euclidean-MDS-stacked}) and diversity space (Figure \ref{fig:diversity-MDS-stacked}).


Remaining experiments will be attached as supplementary material.

\section{Analyzing Diversity and Errors}

\subsection{Average Pairwise Diversity}

\begin{figure}[ht] 
\vskip 0.2in
\begin{center}
\centerline{\includegraphics[width=.75\columnwidth]{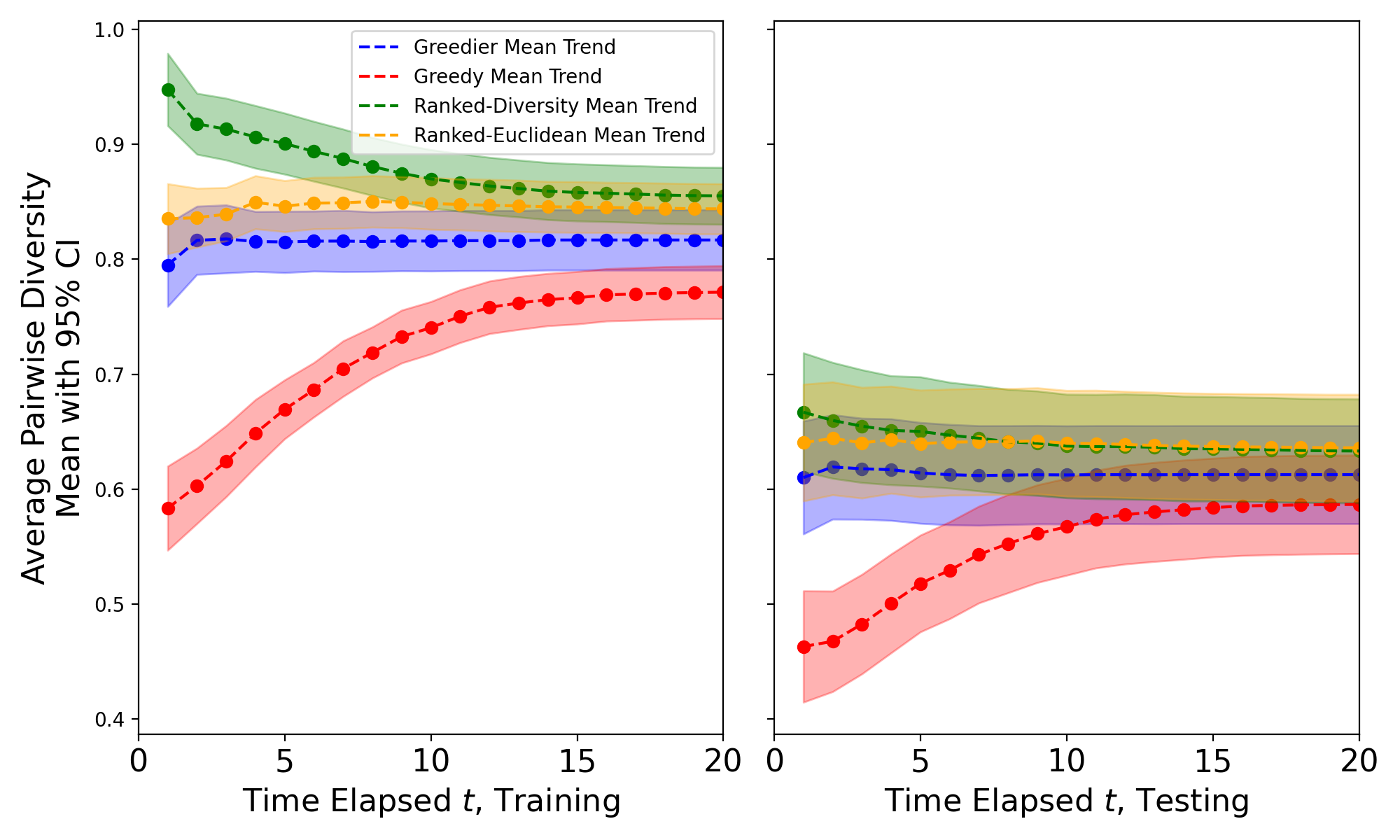}}
\caption{Average pairwise diversity through time for all algorithms across the 40 trials.}
\label{fig:avgpd-all}
\end{center}
\vskip -0.2in
\end{figure}

Given a collection of models $\{\theta_1, \dots, \theta_k \}$ it is unclear how to measure the total diversity of the collection because the ratio-error diversity is defined via pairwise relationships. As such, \cite{rame2022diverse} choose to represent the diversity of the collection as the average pairwise diversity between any two distinct models in the collection. In these algorithms at any time step, we can use the ingredients selected so far to calculate the average pairwise diversity. In Figure \ref{fig:avgpd-all} we plot the average pairwise diversity up to time step 20. 

\begin{figure}[ht] 
\vskip 0.2in
\begin{center}
\centerline{\includegraphics[width=.75\columnwidth]{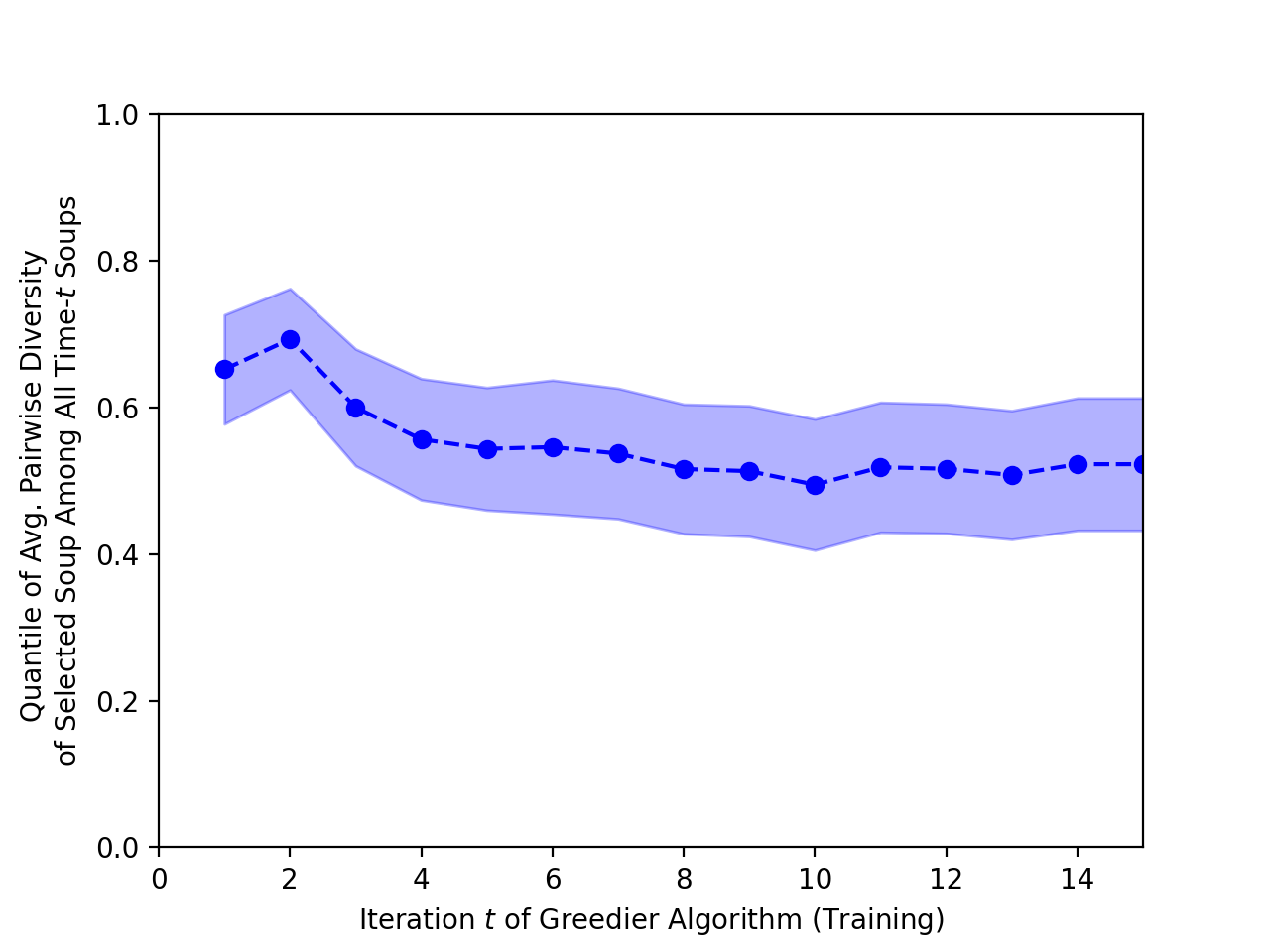}}
\caption{Quantile of average pairwise diversity of selected model through time for all algorithms across the 40 runs.}
\label{fig:avgpd-greedier-quants}
\end{center}
\vskip -0.2in
\end{figure}

In only the greedier algorithm at each time-step do we have access to the result of including each remaining ingredients with the current set. As such, we may calculate the average pairwise diversity of every WA from the time step and bin the quantile of the average pairwise diversity of the selected model. This allows us to see whether WAs with a higher or lower average pairwise diversity are selected. Figure \ref{fig:avgpd-greedier-quants} visualizes the mean trend in the quantile of the new WA at each time step.

\subsection{Dynamics of Errors}

\begin{figure}[ht]
\vskip 0.2in
\begin{center}
\centerline{\includegraphics[width=.75\columnwidth]{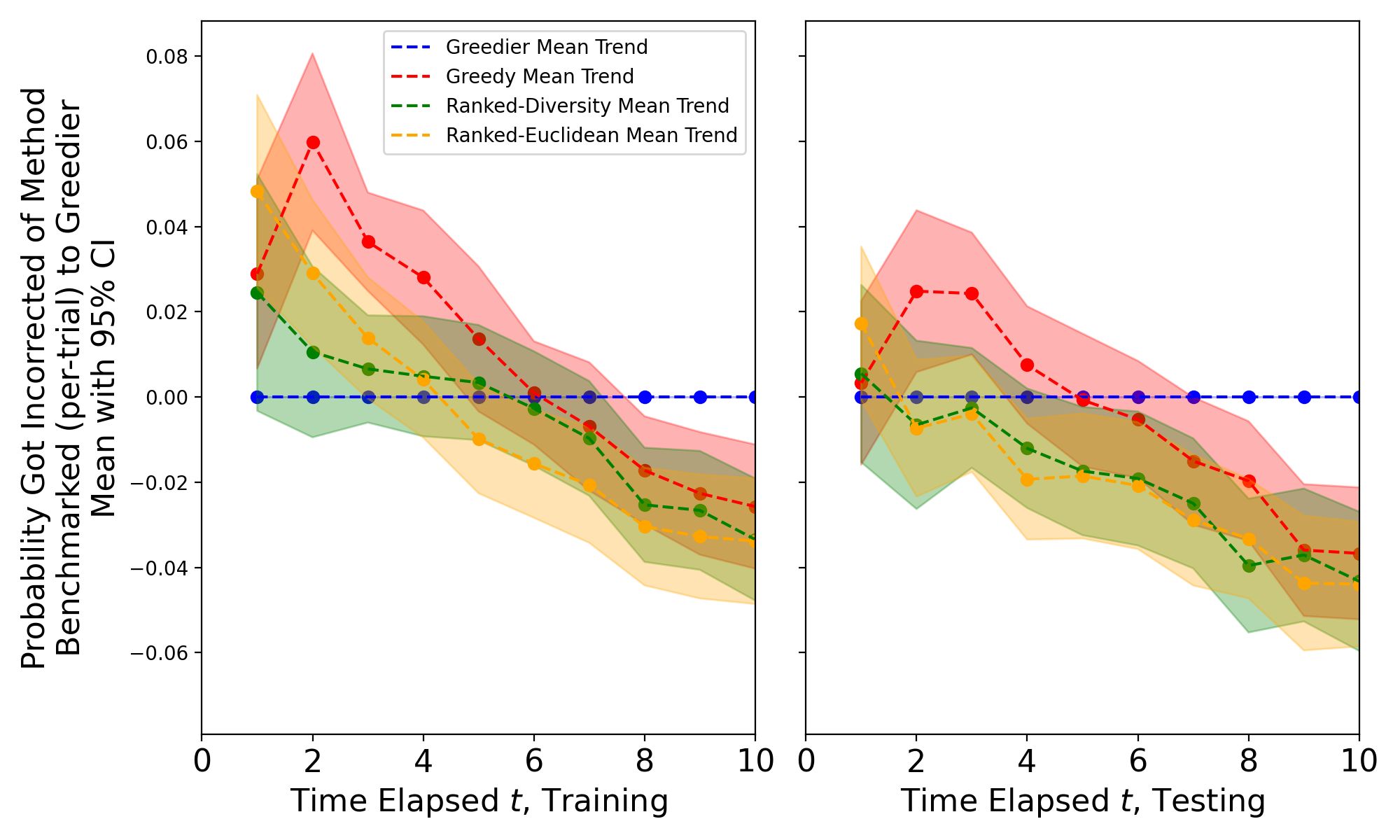}}
\caption{$t$-correct ingredient-incorrect: Series of probabilities that the next-step WA predicts incorrectly given that the current-step WA was correct and the ingredient was incorrect, difference from greedier and other methods' averaged across all trials with 95\% confidence interval. Training at left, testing at right. Terminal value carried forward.}
\label{fig:got-incorrected-10}
\end{center}
\vskip -0.2in
\end{figure}

\begin{figure}[ht]
\vskip 0.2in
\begin{center}
\centerline{\includegraphics[width=.75\columnwidth]{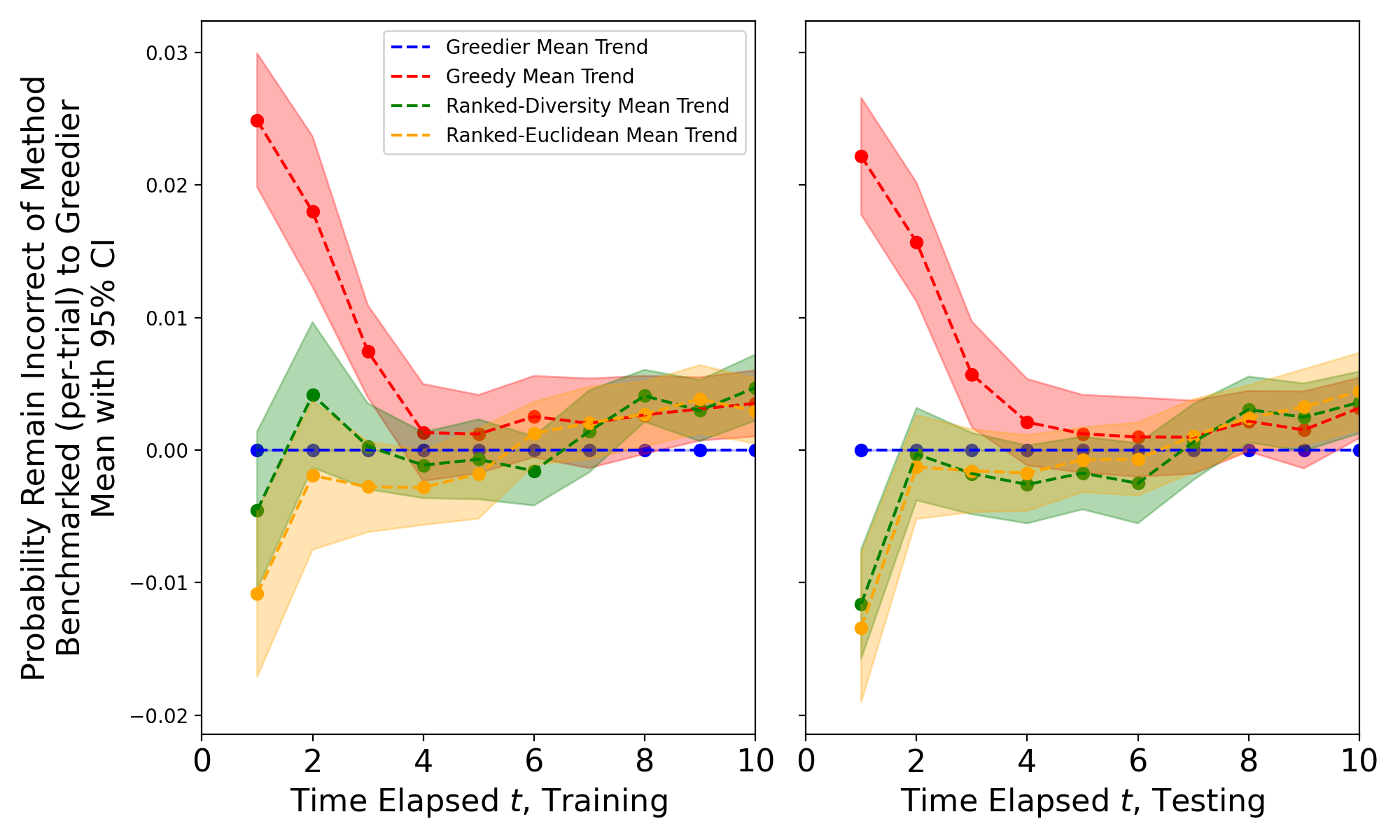}}
\caption{$t$-incorrect ingredient-incorrect: Probabilities that the next-step WA predicts incorrectly given that the current-step WA was incorrect and the ingredient was incorrect, difference from greedier and other methods' averaged across all trials with 95\% confidence interval. Training at left, testing at right. Terminal value carried forward.}
\label{fig:remain-incorrect-10}
\end{center}
\vskip -0.2in
\end{figure}

Similar to the benchmarking of the probability trend in Figure \ref{fig:got-corrected-10}, we plot the results for 2) $t$-correct ingredient-incorrect in Figure \ref{fig:got-incorrected-10}, and 4) $t$-incorrect and ingredient-correct in Figure \ref{fig:remain-incorrect-10}. We do not plot 3) $t$-correct ingredient-correct due to a lack of observable trend. 

From Figure \ref{fig:got-incorrected-10}, we observe in the ID and OOD case for early iterations up to $t = 4$ the greedy method is signifcantly more likely than greedier to make a ``new" error when the ingredient made the error as well. Most of the distribution of values for the diversity-ranked also lie below that of the greedy algorithm, although there is some intersection of the confidence intervals. This evidences that diversity between a WA and an ingredient which makes mistakes may contribute to being more robust to the next-step WA making the same error.

In Figure \ref{fig:remain-incorrect-10}, we observe that the the diversity-selecting methods and the greedier algorithm lower probabilities of retaining a current WA's errors when the ingredient was also correct. This demonstrates that if an ingredient (though well-trained) is incorrect, the WA's performance on previously-misclassified points will benefit more from its inclusion if the model is diverse from the current WA.  

\section{Algorithms}

We formalize our greedier algorithm below as Algorithm \ref{alg:greedier}. Notation for Algorithm \ref{alg:greedier} is drawn from \cite{wortsman2022model}.

\begin{algorithm}[tb]
   \caption{Greedier Algorithm for Weight-Ensembling}
   \label{alg:greedier}
\begin{algorithmic}
   \STATE {\bfseries Input:} Fine-tuned, potential ingredients $\{\theta_1, \dots, \theta_k \}$, sorted by ID validation accuracy in the training domain, and a choice of diversity metric.
   \STATE Initialize ingredients $\leftarrow \{ \theta_1 \}$
   \STATE Initialize remaining ingredients $\leftarrow \{\theta_2, \dots, \theta_j\}$
   \STATE Initialize MaxAcc $\leftarrow \textrm{ValAcc}(\theta_1)$.
   \FOR{$i=2$ {\bfseries to} $k$}
   \STATE $\textrm{Best}_i \leftarrow 0$
   \FOR{$\theta_j$ in remaining ingredients}
   \IF{$\textrm{ValAcc}(\textrm{average}(\textrm{ingredients} \cup \{\theta_j\})) \geq $ MaxAcc}
   \STATE MaxAcc $\leftarrow \textrm{ValAcc}(\textrm{average}(\textrm{ingredients} \cup \{\theta_j\}))$
   \STATE $\textrm{Best}_i \leftarrow j$
   \ENDIF
   \ENDFOR
   \IF{$\textrm{Best}_i > 0$}
   \STATE ingredients $\leftarrow \textrm{ingredients} \cup \{ \theta_j\}$
   \STATE remaining ingredients $\leftarrow \textrm{ remaining ingredients} \setminus \{\theta_j \}$
   \ELSE 
   \STATE \textbf{return} $\textrm{average} (\textrm{ingredients})$ 
   \ENDIF
   \ENDFOR
   
   \STATE \textbf{return} $\textrm{average}(\textrm{ingredients})$

\end{algorithmic}
\end{algorithm}

\begin{algorithm}[tb]
   \caption{Ranked Algorithm for Weight-Ensembling}
   \label{alg:ranked}
\begin{algorithmic}
   \STATE {\bfseries Input:} Fine-tuned, potential ingredients $\{\theta_1, \dots, \theta_k \}$, sorted by ID validation accuracy in the training domain
   \STATE Initialize ingredients $\leftarrow \{ \theta_1 \}$
   \STATE Initialize remaining ingredients $\leftarrow \{\theta_2, \dots, \theta_j\}$
   \STATE Initialize MaxAcc $\leftarrow \textrm{ValAcc}(\theta_1)$.
   \FOR{$i=2$ {\bfseries to} $k$}
   \STATE $\textrm{Best}_i \leftarrow 0$
   \STATE Calculate diversity-metric between $\textrm{average}(\textrm{ingredients} \cup \{\theta_j\})$ and each remaining ingredient
   \FOR{$\theta_j$ Diversity-Ranked-Descending(remaining ingredients)}
        \IF{$\textrm{ValAcc}(\textrm{average}(\textrm{ingredients} \cup \{\theta_j\})) \geq $ MaxAcc}
        \STATE MaxAcc $\leftarrow \textrm{ValAcc}(\textrm{average}(\textrm{ingredients} \cup \{\theta_j\}))$
        \STATE $\textrm{Best}_i \leftarrow j$
        \STATE \textbf{Break loop}
    \ENDIF
    \ENDFOR
   \IF{$\textrm{Best}_i > 0$}
   \STATE ingredients $\leftarrow \textrm{ingredients} \cup \{ \theta_j\}$
   \STATE remaining ingredients $\leftarrow \textrm{ remaining ingredients} \setminus \{\theta_j \}$
   \ELSE 
   \STATE \textbf{return} $\textrm{average} (\textrm{ingredients})$ 
   \ENDIF
   \ENDFOR
   
   \STATE \textbf{return} $\textrm{average}(\textrm{ingredients})$

\end{algorithmic}
\end{algorithm}

\end{document}